\documentclass{article}

\usepackage[preprint]{neurips_2026}

\usepackage[utf8]{inputenc} % allow utf-8 input
\usepackage[T1]{fontenc}    % use 8-bit T1 fonts
\usepackage{hyperref}       % hyperlinks
\usepackage{url}            % simple URL typesetting
\usepackage{booktabs}       % professional-quality tables
\usepackage{amsfonts}       % blackboard math symbols
\usepackage{nicefrac}       % compact symbols for 1/2, etc.
\usepackage{microtype}      % microtypography
\usepackage{xcolor}         % colors
\usepackage{amsmath}
\usepackage{listings}
\usepackage{enumitem}
\usepackage{multicol}
\usepackage{multirow}
\usepackage{wrapfig}
\usepackage{tabularx}
\usepackage{array}
\usepackage{tcolorbox}
\tcbuselibrary{listings,breakable}
\definecolor{amethyst}{rgb}{0.6, 0.4, 0.8}

\definecolor{lemon}{RGB}{255,247,0}
\definecolor{maize}{RGB}{250,237,94}
\definecolor{mustard}{RGB}{255,219,89}
\definecolor{ocre}{RGB}{241,103,35}
\definecolor{Tangerine}{RGB}{253,128,8}
\definecolor{framegreen}{RGB}{153, 188, 133}
\definecolor{bggreen}{RGB}{235, 250, 228}
\definecolor{c0}{cmyk}{1,0.3968,0,0.2588} 
\definecolor{c1}{cmyk}{0,0.6175,0.8848,0.1490} 
\definecolor{c2}{cmyk}{0.1127,0.6690,0,0.4431} 
\definecolor{c3}{cmyk}{0.3081,0,0.7209,0.3255} 
\definecolor{c4}{RGB}{164, 16, 52}
\definecolor{orange}{HTML}{E66100}
\definecolor{bluex}{HTML}{0C7BDC}
\definecolor{yellow}{HTML}{FFC20A}
\definecolor{lightpurple}{HTML}{E6E6FA}
\definecolor{lightbluee}{HTML}{e8f4f8}
\definecolor{blush}{rgb}{0.87, 0.36, 0.51}
\definecolor{c5}{HTML}{EE4E4E}
\definecolor{gggggg}{HTML}{EFEFEF}
\ifnum1=1

\usepackage{inconsolata}
\usepackage[T1]{fontenc}
\usepackage[scaled=0.95]{biolinum}

\usepackage{tcolorbox}
\definecolor{chart}{HTML}{1f77b4}

\newtcolorbox{example}[1][]{
  colback=chart!5!white,
  colframe=chart,
  floatplacement=floating,
  title=\centering \textsf{\small #1}
}
\newtcbox{\hlprimarytab}{on line, box align=base, colback=BlueGreen!20,colframe=blue,size=fbox,arc=3pt, before upper=\strut, top=-2.5pt, bottom=-4.5pt, left=-2pt, right=-2pt, boxrule=0pt}
\newtcbox{\hlsecondarytab}{on line, box align=base, colback=WildStrawberry!10,colframe=orange,size=fbox,arc=3pt, before upper=\strut, top=-2.5pt, bottom=-4.5pt, left=-2pt, right=-2pt, boxrule=0pt}
\newtcbox{\hlwhite}{on line, box align=base, colback=WildStrawberry!8,colframe=white,size=fbox,arc=2pt, before upper=\strut, top=-3pt, bottom=-4.5pt, left=-2pt, right=-2pt, boxrule=0pt}
\newtcbox{\hlyellow}{on line, box align=base, colback=BlueGreen!10,colframe=white,size=fbox,arc=2pt, before upper=\strut, top=-3pt, bottom=-4.5pt, left=-2pt, right=-2pt, boxrule=0pt}

\setlist[itemize]{
    leftmargin=1.5em,  % 设置左边距，可以根据需要调整数值
    rightmargin=0pt, % 设置右边距，0pt 表示不额外缩进
    itemsep=0.25em,     % 项目之间的垂直间距 (可选)
    topsep=0.5em       % 列表环境与上方文本的间距 (可选)
}

\title{Uncovering Entity Identity Confusion \\ in Multimodal Knowledge Editing}

\author{%
Shu Wu\textsuperscript{1}\thanks{Major Contributors.}  ,
Xiaotian Ye\textsuperscript{2}$^*$,
Xinyu Mou\textsuperscript{1,3}$^*$,
Dongsheng Liu\textsuperscript{1,4}$^*$,
Xiaohan Wang\textsuperscript{5},
Mengqi Zhang\textsuperscript{6}\thanks{Corresponding author.}
\\ \\
\textsuperscript{1}New Laboratory of Pattern Recognition (NLPR) \\
State Key Laboratory of Multimodal Artificial Intelligence Systems (MAIS)\\
Institute of Automation, Chinese Academy of Sciences\\
\textsuperscript{2}Beijing University of Posts and Telecommunications\\
\textsuperscript{3}School of Artificial Intelligence, University of Chinese Academy of Sciences\\
\textsuperscript{4}School of Advanced Interdisciplinary Sciences, University of Chinese Academy of Sciences\\
\textsuperscript{5}Huazhong University of Science and Technology\\
\textsuperscript{6}Shandong University
\\
\texttt{shu.wu@nlpr.ia.ac.cn, yexiaotian@bupt.edu.cn}\\
\texttt{\{mouxinyu2025, liudongsheng2025\}@ia.ac.cn}\\
\texttt{shawn\_wang@hust.edu.cn, mengqi.zhang@sdu.edu.cn}
}

\begin{document}

\maketitle

\begin{abstract} 
Multimodal knowledge editing (MKE) aims to correct the internal knowledge of large vision-language models after deployment, yet the behavioral patterns of post-edit models remain underexplored. In this paper, we identify a systemic failure mode in edited models, termed Entity Identity Confusion (EIC): edited models exhibit an absurd behavior where text-only queries about the original entity's identity unexpectedly return information about the new entity. To rigorously investigate EIC, we construct EC-Bench, a diagnostic benchmark that directly probes how image-entity bindings shift before and after editing. Our analysis reveals that EIC stems from existing methods failing to distinguish between Image-Entity (I-E) binding and Entity-Entity (E-E) relational knowledge in the model, causing models to overfit E-E associations as a shortcut: the image is still perceived as the original entity, with the new entity's name serving only as a spurious identity label. We further explore potential mitigation strategies, showing that constraining edits to the model's I-E processing stage encourages edits to act more faithfully on I-E binding, thereby substantially reducing EIC. Based on these findings, we discuss principled desiderata for faithful MKE and provide methodological guidance for future research.
\end{abstract}

\section{Introduction}

Today's knowledge editing (KE) \citep{zhang2024comprehensivestudyknowledgeediting} has established itself as a key research area in the large language model (LLM) \citep{zhao2025surveylargelanguagemodels} field. In real-world deployments, maintaining LLMs often requires revising their encoded knowledge to address outdated facts or to meet safety, policy, and privacy requirements. Knowledge editing focuses on targeted modifications to the internal knowledge of LLMs, thereby enabling more practical and auditable post-deployment maintenance. With the growing adoption of large vision-language models (LVLMs) \citep{liu2023llava, zhu2023minigpt, Qwen-VL} in real-world applications, these needs have naturally extended from purely textual systems to \textbf{Multimodal Knowledge Editing (MKE)} \citep{cheng-etal-2023-edit}.

\begin{figure}
    \centering
    \includegraphics[width=1\linewidth]{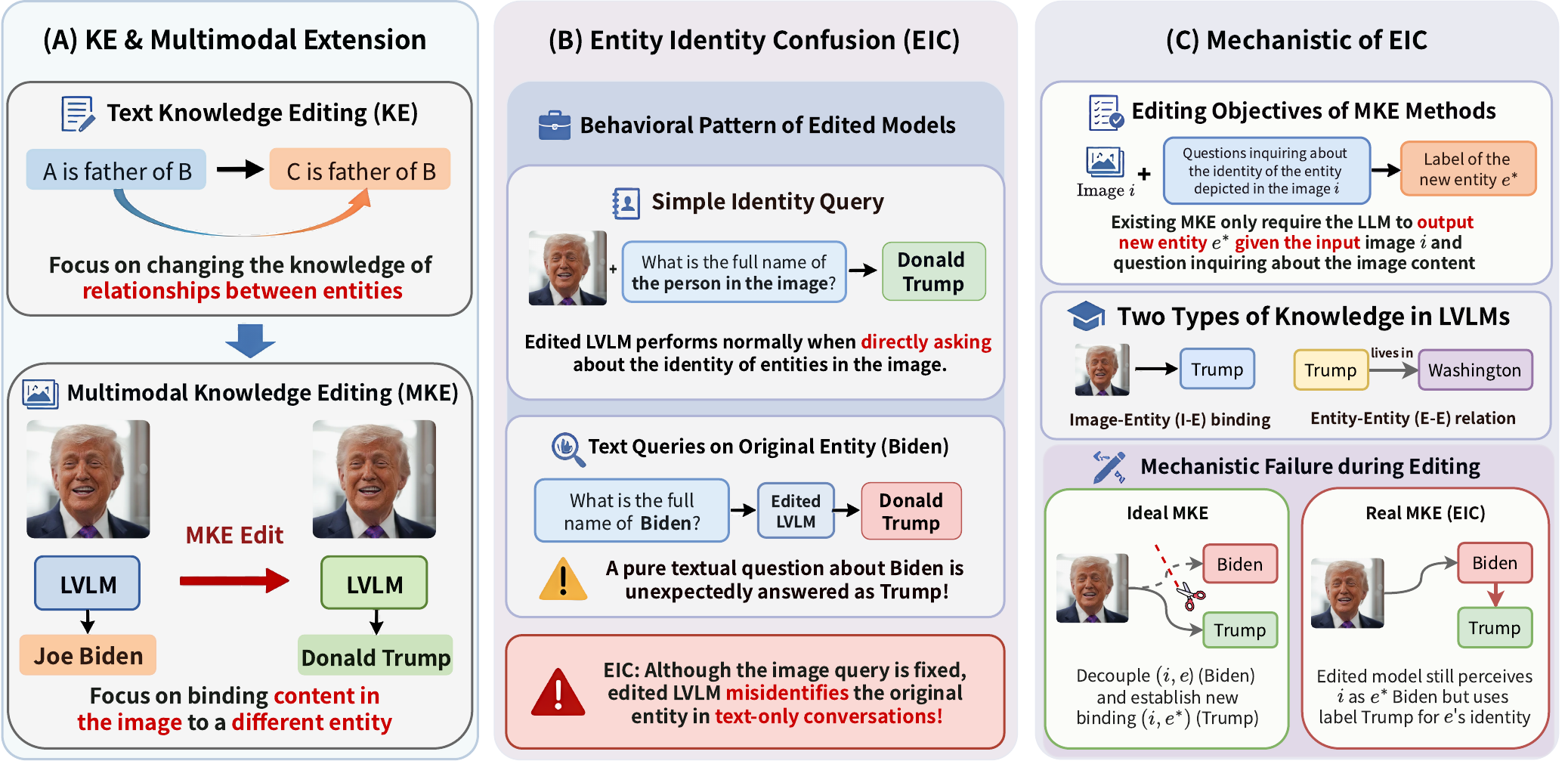}
    \caption{Overview of Entity Identity Confusion (EIC) in multimodal knowledge editing.}
    \label{fig:main}
    \vspace{-2em}
\end{figure}

Unlike text-based knowledge editing \citep{meng2022locating, zhang2026spectralcharacterizationmitigationsequential},  which typically targets relationships between real-world entities (e.g., modifying that ``\textit{Trump, graduate from, UPenn}''), mainstream multimodal KE settings focus on binding the content depicted in a specific image to a different entity. As shown in Figure \ref{fig:main}(a), for an image $A$ of Trump that the pre-edit model erroneously recognizes as Biden, the post-MKE model correctly identifies the content in the image as the true entity Trump. Despite this natural motivation, multimodal KE remains considerably less mature than its text-only counterpart, and systematic analysis of post-edit model behavior is largely absent from the literature.

In this work, we observe a previously undiscovered failure mode during our analysis of post-edit model behavior, which we term \textbf{Entity Identity Confusion (EIC)}: after the entity bound to image $i$ is modified from $e$ to $e^*$, when asked identity-related questions about $e$, the model surprisingly responds with the name of $e^*$. To illustrate this issue, consider the aforementioned case of rectifying the image-entity association for Trump: as illustrated in Figure \ref{fig:main}(b), when prompted with identity queries such as ``\textit{Who is this?}'', the edited model may indeed output ``\textit{Trump},'' and its performance might appear normal under existing benchmark metrics. However, deeper probing reveals a behavior that even non-experts would find absurd: when the model is asked text-only questions about \textit{Biden} (the entity previously associated with the image before editing), such as ``\textit{What is the full name of Biden?}'', the model unexpectedly answers ``\textit{Trump}'' This is clearly highly anomalous. We conducted a pilot study and consistently observed this pattern across various editing methods, indicating that such an issue is a systemic phenomenon rather than an isolated error.

We further perform an in-depth analysis of the characteristics of EIC. Given that EIC is difficult to detect using standard metrics in traditional benchmarks, we construct a more comprehensive benchmark, \textbf{EC-Bench}. In addition to tasks specifically designed to examine EIC, EC-Bench introduces two generalization tasks: Old Binding Persistence (OBP), and New Binding Generalization (NBG), to evaluate how the bindings between images and the original/new entities evolve after editing. This allows us to analyze more characteristics of EIC and explore its underlying mechanisms. Ideally, MKE should decouple image $i$ from the original entity $e$ and establish a new binding with entity $e^*$. Our experimental analysis, however, reveals that existing MKE methods largely fail to affect the image-entity binding; instead, the edited model still perceives $i$ as the original entity $e$ (e.g., Biden) but uses the $e^*$ label ``\textit{Trump}'' to describe 
$e$’s identity, which explains the phenomena we observed. Consequently, on more complex tasks such as asking ``\textit{Which university did the person in the image graduate from?}'', the model still provides the alma mater of Biden. This suggests that even when the internal mechanism is fundamentally flawed, the model can still exhibit seemingly ideal behavior on simple tasks, thereby ``deceiving'' many existing benchmarks.

What causes EIC? We posit that EIC stems from the fact that existing MKE methods fail to explicitly account for the complexity of different knowledge types in multimodal settings. As shown in Figure \ref{fig:main}(c), the objectives of current MKE methods typically only require the model to produce the correct string on given samples \citep{NEURIPS2024VLKEB}, a superficial behavioral constraint: they achieve this through parameter updates and similar mechanisms, without any constraint of how it is internally realized. However, knowledge in LVLMs involves two distinct categories \citep{zhang-etal-2025-mc}: Image-Entity (I-E) binding  $(i, e)$  and Entity-Entity (E-E) relations $(  e_{1}, r, e_{2}  )$, which may rely on different retrieval mechanisms at the model’s architecture levels. \textbf{This discrepancy means the model may in practice satisfy the editing objective through incorrect underlying mechanisms.} For instance, the model may implicitly force a spurious association between Biden and Trump -- which yields correct answers on simple questions but is fundamentally incorrect at the underlying level, exposing issues like EIC under complex tests.

We therefore advocate that a principled editing strategy should decouple two types of knowledge, ensuring that editing interventions precisely target I-E binding representations while preserving the structural integrity of E-E relational knowledge. To provide methodological guidance for future research, we further we further explored and proposed a potential mitigation strategy for EIC: we propose that, since I-E recall and E-E recall occur at different locations during model inference, \textbf{restricting the editing target to the region responsible for I-E binding may help direct the editing effect toward the correct type of knowledge}, thereby mitigating EIC and enabling more accurate knowledge editing. We validate this hypothesis across multiple baseline methods by varying the editing location, and confirm that this constitutes a promising and robust direction for future research. Furthermore, we discuss future directions for correct multimodal knowledge editing, thereby providing principled guidance for future MKE research.

% We further explore potential mitigation strategies for EIC to provide actionable context for future research. In evaluating existing baselines on EC-Bench, we observe a notable finding: FT-Vis, an editing strategy that targets the vision encoder rather than language components, exhibits substantially reduced EIC. This is intuitive as the vision-side parameters cannot encode E-E relational knowledge by definition, so edits applied there are unlikely to inadvertently corrupt E-E associations through overfitting. This observation motivates our hypothesis that, since I-E and E-E recall occur at different locations during model inference, constraining the editing target to regions responsible for I-E binding may encourage edits to act on the correct knowledge type, thereby mitigating EIC and enabling more faithful knowledge editing. We test this hypothesis across multiple baseline methods by varying the editing locus and confirm that this constitutes a promising and robust direction for future work. We further provide a detailed discussion of the desiderata for correct multimodal knowledge editing -- articulating what a properly executed MKE should achieve in terms of selectively updating I-E bindings without corrupting E-E knowledge—thereby offering principled guidance for future MKE research.

The core contributions of this paper are summarized as follows:

\begin{itemize}
\item We identify and define Entity Identity Confusion (EIC) as an overlooked systematic failure mode in multimodal knowledge editing.
\item We construct a diagnostic benchmark EC-Bench and introduce more demanding generalization tasks to thoroughly assess the internal knowledge structure of the edited model, facilitating future in-depth analysis of this issue.
\item We conduct mechanistic diagnosis and analysis of MKE based on the benchmark, and propose a preliminary mitigation strategy, thereby providing methodological guidance for future multimodal editing research.
\end{itemize}

\section{Preliminaries}

This section provides definitions of key concepts and necessary backgrounds relevant to our work.

\subsection{Architecture of Large Vision-Language Models}

A typical large vision-language model (LVLM) \citep{liu2023llava, zhu2023minigpt, li2023blip2} consists of three components: a \textbf{vision encoder}, a \textbf{projector}, and an \textbf{LLM backbone}. 

Given an input image $i$, the vision encoder (e.g., a Vision Transformer) extracts a sequence of visual token embeddings $\mathbf{v} = [v_1, \dots, v_n]$. The projector (e.g., a linear layer or MLP) maps these tokens into the LLM's embedding space, yielding $\mathbf{h} = \mathrm{Proj}(\mathbf{v})$. The LLM backbone then takes the concatenation of $\mathbf{h}$ and the text token embeddings as input and performs autoregressive generation to produce the output.

\subsection{Problem Formulation}

\textbf{Knowledge in LVLMs} can be decomposed into two distinct types \citep{zhang-etal-2025-mc}. \textit{Image-entity (I-E) binding knowledge} $(i,e)$ captures the correspondence between visual evidence and entity identity, answers ``who or what does this image refer to?'' \textit{Entity-entity (E-E) relational knowledge} $(e_1, r, e_2)$ captures facts and attributes connected to an entity through semantic relations, such as birthplace, occupation, or affiliation. These two types may be handled by different components and layers of the model, a premise that motivates our analysis in later sections.

\textbf{Multimodal Knowledge Editing (MKE)} aims to modify I-E bindings: given an image $i$ originally bound to entity $e$, the goal is to rebind it to a target entity $e^*$. Formally, let $f(\cdot;\theta)$ denote a pretrained LVLM with parameters $\theta$. Given an image $i$ and a textual query $x$, the model outputs an answer $y=f(i,x;\theta)$. We are given an edit set
\begin{equation}
\mathcal{D}_{\text{edit}} = \{(i, x, y, y')\},
\end{equation}
where $x$ is a query about the identity of the entity depicted in $i$, $y$ is the model-consistent pre-edit answer, and $y'$ is the target answer expected after editing. 

An editing method $\mathcal{M}$ produces updated parameters $\theta' = \mathcal{M}(\theta, \mathcal{D}_{\text{edit}})$. The standard objective is
\begin{equation}
f(i, x; \theta') = y',
\end{equation}
while preserving unrelated model behavior.

\section{Observing Entity Identity Confusion: A Preliminary Experiment}

To empirically validate Entity Identity Confusion (EIC), 
% where the edited model produces highly anomalous outputs regarding the entities described in the original image even on the pure text modality, 
we conduct a preliminary experiment. In this section, We first detail the experimental setup, including the evaluation tasks we adopt. Subsequently, based on the experimental results, we elaborate on the performance of EIC in downstream tasks and verify its prevalence across different basemodels and MKE methods.

\subsection{Preliminary Experiments Settings}

Our preliminary experiments are based on a representative MKE Benchmark, VLKEB \citep{NEURIPS2024VLKEB}, and extend its pipeline with additional evaluation tasks targeting EIC to observe the post-edit behavior of models under various editing methods. Descriptions of the baselines are provided in Appendix \ref{apd:baselines}.

\textbf{Editing Task.} The editing objective of MKE is to modify an image-entity binding within the model, i.e., $(i,e) \rightarrow (i,e')$. In practice, it provides a set of training samples containing images paired with questions querying the identity of the entity depicted; for example, \texttt{[Image of Biden] What's the full name of the person in this image?}; and requires performing a \textbf{counterfactual} edit such that the model responds with \texttt{Donald Trump}.

\textbf{Evaluation Task.}  To evaluate EIC, we query the identity of the original entity $e$ in a pure text modality that contains no images, and examine the proportion of cases where the model erroneously predicts the label of the new entity, $e^*$, as the answer. For example, we ask \textit{What's the full name of Biden?} Models exhibiting EIC will anomalously respond with \textit{Donald Trump}. We also provide the efficacy metric, which is the classic edit success rate metric.

\subsection{Characteristics of EIC}

% 应该有个柱状图 放不同方法的edit success & EIC指标

We observe three recurring characteristics of EIC from the preliminary experiment.
% \textbf{Characteristic 1: High Efficacy, High Confusion.} Across methods, direct edit success can be high while identity confusion remains substantial. This disconnect implies that single-prompt efficacy is insufficient as a sole indicator of edit quality in LVLMs. In other words, passing the edited prompt does not imply faithful identity rewiring.

\begin{wrapfigure}{r}{0.45\textwidth}
\vspace{-1.4em}
\centering
\includegraphics[width=\linewidth]{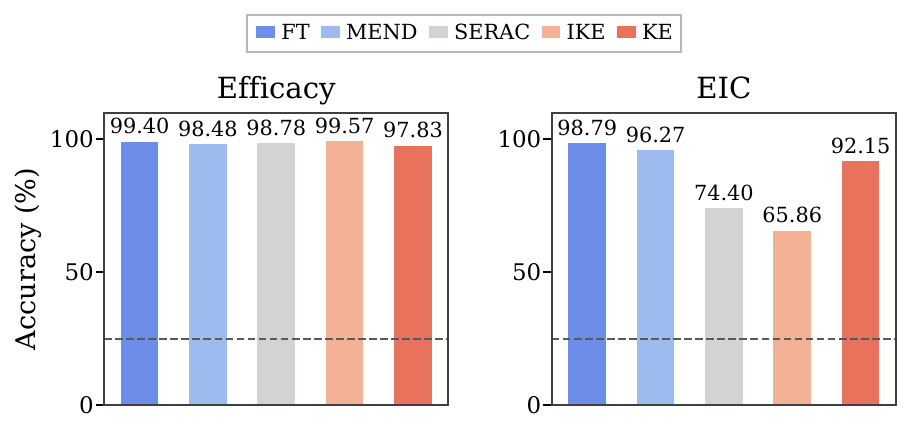}
\caption{\small Performance of LLaVA edited with various MKE methods.}
\label{fig:prelim-eic-case}
\vspace{-0.8em}
\end{wrapfigure}

% \textbf{Characteristic 2: Persistence Across Method Families.} EIC appears in parameter-editing methods, memory-based methods, and even prompt-based editing variants. The exact severity varies, but the pattern itself is recurrent. This suggests that EIC is not simply a bug of one implementation; it is a structural risk induced by how current methods couple multimodal grounding with language-space edits.

% \textbf{Characteristic 3: Cross-Modal Spillover.} A visually triggered edit can alter text-only identity behavior. This indicates strong cross-modal entanglement of entity representations: changing a local image-conditioned response can affect global textual identity channels. Such spillover is especially problematic for applications requiring stable textual knowledge while performing targeted multimodal updates.

\textbf{Characteristic 1: High Efficacy Coexists with High Confusion.} Across all editing methods, models achieve high edit success rates on the original edit queries while simultaneously exhibiting severe identity confusion. This implies that single-prompt efficacy is insufficient as a sole indicator of edit quality in LVLMs.

\textbf{Characteristic 2: Universality Across Editing Paradigms.} EIC is not confined to any single class of editing methods. It manifests in parameter-modifying approaches (e.g., FT, MEND), external-memory-based methods (e.g., SERAC), and prompt-based strategies (e.g., IKE) alike. While the severity differs across methods, the recurrence of this pattern across fundamentally different editing paradigms indicates that EIC is a structural issue inherent to the current MKE formulation.

\textbf{Characteristic 3: Text-side Knowledge Contamination.} MKE targets the model's I-E binding, which should be image-conditioned behavior that only manifests when image input is provided; however, we observe that the model also exhibits clearly anomalous behavioral patterns under text-only queries, indicating that the editing has contaminated the model's textual knowledge representations rather than acting precisely on the I-E relationship.

\textbf{Conclusion.} Based on these observations, we provide a formal definition of the EIC phenomenon. Given an editing instance that rebinds image $i$ from entity $e$ to target entity $e^*$, we define EIC as the phenomenon where the post-edit model $f(\cdot;\theta')$, when queried about the identity of $e$ through a text-only prompt $x_{\text{text}}$ (i.e., without any image input), erroneously outputs $e^*$:
\begin{equation}
\text{EIC}: \quad f(x_{\text{text}}^{(e)}; \theta') = e^*, \quad \text{where } f(x_{\text{text}}^{(e)}; \theta) = e.
\end{equation}
In other words, the editing procedure intended to modify only the correspondence between images and entities, which is visual-conditioned behavior, but causes the model to conflate the identities of $e$ and $e^*$ even in the absence of any visual input.

% Taken together, these findings motivate the need for a dedicated confusion-aware evaluation framework. In the next section, we formalize this need by defining EC-Bench tasks that separately quantify identity corruption, old-binding persistence, and new-binding generalization, thereby exposing failure dimensions that standard efficacy-centric protocols cannot capture.

\section{Analyzing Post-Edit Binding Behavior with EC-Bench}

To provide a more detailed analysis of how EIC manifests across different model architectures and editing methods, we introduce \textbf{EC-Bench} (Entity Confusion Benchmark), an evaluation framework that extends standard MKE protocols \citep{NEURIPS2024VLKEB, cheng-etal-2023-edit} with dedicated diagnostics for identity corruption and binding inconsistency. In this section, we first describe the tasks introduced by EC-Bench, and then assess the performance of editing methods, accompanied by a diagnostic analysis of how internal knowledge associations are altered in post-edit models.

\subsection{EC-Bench}

EC-Bench consists of three \textbf{fundamental tasks} and three \textbf{binding diagnostic tasks}. The fundamental tasks align with conventional MKE benchmark settings and measure each method's basic editing competency, covering \textbf{Efficacy, Generality, and Locality}. The binding diagnostic tasks are specifically designed to detect the EIC phenomenon and to analyze how internal knowledge associations are formed in edited models; to this end, we introduce three dedicated probes: \textbf{Entity Identity Confusion (EIC), Old Binding Persistence (OBP)}, and \textbf{New Binding Generalization (NBG)}.

\textbf{Fundamental Tasks.}  Specifically, we introduce the following three fundamental tasks.
\begin{itemize}
    \item \textbf{Efficacy} measures whether the edited model returns target entity $e^*$ on the original edit query. This is the minimal criterion for successful intervention.

    \item \textbf{Generality} evaluates whether edited behavior transfers to semantically equivalent variants. \emph{T-Gen} uses paraphrased text prompts with the same image; \emph{I-Gen} uses alternative images of the same entity with the same query intent. High generality indicates that the edit is not merely a string-level patch to one prompt template.

    \item \textbf{Locality} measures whether unrelated knowledge remains stable. \emph{T-Loc} compares pre-/post-edit answers on unrelated text-only queries; \emph{I-Loc} compares pre-/post-edit behavior on visually similar but non-target entities.
\end{itemize}
\textbf{Binding Diagnostic Tasks.}
Consider the running example where an image $i$ of \textit{Biden} ($e$) is edited to be rebound to \textit{Trump} ($e^*$). If we use a multimodal knowledge graph \citep{liu2019mmkg} to represent the underlying knowledge structure of the model, MKE is primarily concerned with three edges: (1) avoid introducing a spurious E-E edge $(\text{Biden}, \text{Trump})$, (2) erase the old I-E edge $(i, \text{Biden})$, and (3) establish the new I-E edge $(i, \text{Trump})$. We introduce three binding diagnostic tasks to probe these three edges respectively, thereby characterizing how editing alters entity binding at a finer granularity.

\begin{itemize}
    \item \textbf{Entity Identity Confusion (EIC)} probes edge~(1): whether a spurious E-E association $(e, e^*)$ has been created. After editing, we ask identity questions about $e$ without image input (e.g., \textit{What is the full name of Biden?}). If the model responds with $e^*$ (\textit{Trump}), we count it as confusion.

    \item \textbf{Old Binding Persistence (OBP)} probes edge~(2): whether the old I-E binding $(i,e)$ still survives after editing. Note that directly asking ``\textit{Who is in this image?}'' cannot reliably test this, because the spurious E-E edge from EIC may redirect the answer to $e^*$ even when the model still internally perceives $i$ as $e$. We therefore test the old binding \emph{indirectly} via multi-hop reasoning $(i \rightarrow e, r, e_1)$: we present image $i$ and ask relational facts unique to $e$ (e.g., ``\textit{Which university did the person in this image graduate from?}''). Correct answers for $e$ indicate the old binding remains active. 

    \item \textbf{New Binding Generalization (NBG)} probes edge~(3): whether the new binding $(i,e^*)$ supports factual reasoning beyond the edited prompt. This task takes the form of a multi-hop reasoning task consistent with OBP, but probes relations involving the new entity $(i \rightarrow e^*, r, e_2)$: we present image $i$ and query facts unique to $e^*$ (e.g., ``\textit{In which city was the person in this image born?}''). Correct answers for $e^*$ indicate that the model has formed a functional new grounding rather than merely memorizing one output string.
\end{itemize}

\subsection{Experiments and Findings}
\label{subsec:main-expr}

To conduct a thorough analysis of EIC, we employ six editing methods: FT-Vis, FT-LLM, KE, MEND, IKE, and SERAC (Details in Appendix.\ref{apd:baselines}), to edit LLaVA-1.5 \citep{liu2023llava}, MiniGPT-4 \citep{zhu2023minigpt}, mPLUG-Owl2 \citep{ye2023mplugowl2}, and Qwen-VL \citep{Qwen-VL}, evaluating performance on EC-Bench. Detailed results are presented in Table~\ref{tab:main}, while results for Owl2 are presented in Appendix \ref{apd:owl-results}. Based on these results, we summarize our findings as follows:

% Auto-generated from whitelist-backed results.
\begin{table*}[t]
\centering
\footnotesize
\setlength{\tabcolsep}{3.1pt}
\renewcommand{\arraystretch}{1.10}

\caption{Main EC-Bench results on inherited and diagnostic metrics.}
\label{tab:main} 
\sf
\resizebox{\textwidth}{!}{%

\begin{tabular}{@{}cccccccccc@{}}
\toprule
Model & Method & Efficacy $\uparrow$ & T-Gen $\uparrow$ & I-Gen $\uparrow$ & T-Loc $\uparrow$ & I-Loc $\uparrow$ & EIC $\downarrow$ & OBP $\downarrow$ & NBG $\uparrow$ \\
\midrule
\multirow[c]{7}{*}{\textbf{LLaVA}} & \textit{base (unedited)} & 32.4 & 35.8 & 32.1 & 100.0 & 100.0 & 26.1 & 92.2 & 40.4 \\
\cmidrule(lr){2-10}
 & FT & 99.4 & 99.1 & 99.4 & 59.3 & 24.3 & 99.1 & 72.3 & 33.7 \\
 & FT-VIS & 99.9 & 98.1 & 96.4 & 100.0 & 18.8 & 26.1 & 51.2 & 54.6 \\
% \cmidrule(lr){2-10}
 & MEND & 99.0 & 98.7 & 98.9 & 81.4 & 86.8 & 97.2 & 92.1 & 38.8 \\
 & SERAC & 99.6 & 98.6 & 99.6 & 43.5 & 2.0 & 75.4 & 92.2 & 48.1 \\
 & IKE & 100.0 & 99.6 & 100.0 & 53.0 & 14.4 & 66.9 & 45.4 & 63.2 \\
 & KE & 99.2 & 98.5 & 99.0 & 61.7 & 16.3 & 96.1 & 92.7 & 38.1 \\
 
\midrule

\multirow[c]{7}{*}{\textbf{MiniGPT-4}} & \textit{base (unedited)} & 29.4 & 32.1 & 29.1 & 100.0 & 100.0 & 31.8 & 73.0 & 38.1 \\
\cmidrule(lr){2-10}
 & FT & 100.0 & 99.5 & 99.9 & 83.2 & 47.1 & 69.7 & 51.0 & 38.9 \\
 & FT-VIS & 100.0 & 99.5 & 100.0 & 100.0 & 19.8 & 31.8 & 65.5 & 39.5 \\
% \cmidrule(lr){2-10}
 & MEND & 99.4 & 99.1 & 99.3 & 82.6 & 85.5 & 94.2 & 76.9 & 37.6 \\
 & SERAC & 98.9 & 96.2 & 98.8 & 36.0 & 3.8 & 80.2 & 72.2 & 50.2 \\
 & IKE & 100.0 & 99.6 & 100.0 & 61.9 & 9.5 & 67.5 & 70.6 & 50.1 \\
 & KE & 99.2 & 98.9 & 99.1 & 55.1 & 15.7 & 89.9 & 25.7 & 37.0 \\
 
\midrule

\multirow[c]{7}{*}{\textbf{Qwen-VL}} & \textit{base (unedited)} & 25.4 & 28.6 & 25.4 & 100.0 & 100.0 & 24.7 & 72.6 & 31.7 \\
\cmidrule(lr){2-10}
 & FT & 100.0 & 99.0 & 99.9 & 60.8 & 26.8 & 88.2 & 54.0 & 30.7 \\
 & FT-VIS & 100.0 & 96.0 & 99.8 & 100.0 & 4.0 & 24.7 & 35.0 & 32.1 \\
% \cmidrule(lr){2-10}
 & MEND & 99.5 & 98.5 & 98.1 & 86.4 & 86.5 & 68.8 & 72.1 & 32.1 \\
 & SERAC & 82.9 & 80.1 & 83.3 & 29.3 & 29.1 & 53.6 & 57.2 & 36.1 \\
 & IKE & 99.5 & 99.2 & 99.5 & 33.8 & 10.5 & 57.5 & 25.4 & 55.5 \\
 & KE & 99.6 & 96.4 & 99.2 & 39.8 & 19.9 & 92.3 & 49.6 & 32.1 \\
\bottomrule

\end{tabular}}

\end{table*}

\textbf{Finding 1. Nearly all editing methods exhibit severe EIC.} As shown in Table~\ref{tab:main}, every method produces a significant and anomalous increase in EIC scores relative to the base model. FT and MEND on LLaVA even reach a confusion rate approaching 99\%, and the phenomenon is pervasive across different LLM backbones. Such high rates reveal that existing methods cause severe contamination of textual-modal knowledge when editing I-E bindings: even under purely text-based queries, the post-edit model produces highly erroneous outputs with extremely high probability. This clearly violates the expectations for knowledge editing in real-world deployment.

\textbf{Finding 2. Results on challenging tasks reveal that existing editing methods fail to achieve their underlying editing objectives.} A successful MKE intervention should dissolve the binding $(i,e)$ and establish a new $(i,e^*)$. These two core objectives are measured by the OBP and NBG tasks, respectively. However, as shown in Table~\ref{tab:main}, performance on both metrics remains far from satisfactory: post-edit models still retain very high OBP scores, with methods such as MEND and SERAC yielding values that remain close to those of the pre-edit baseline; on the NBG task, the majority of models still score very low, indicating that it is extremely difficult for models to leverage the I-E binding injected during editing for complex reasoning. Overall, NBG scores are consistently and substantially lower than OBP scores, suggesting that the model's internal processing pipeline still tends to first recognize the image as the original entity before performing downstream reasoning.

\textbf{Finding 3. Methods that edit the visual side of models exhibit less EIC, though they still fall short on OBP and NBG.} Among the baseline methods compared in the main experiment, there is a category of approaches that perform editing on the visual side: FT-Vis, which targets the vision encoder or projector module of LVLMs. As shown in Table~\ref{tab:main}, FT-Vis achieves the best EIC scores among all compared methods, approaching the performance of the unedited base model, indicating that it barely contaminates the model's purely text-modal knowledge during the editing process. We attribute this to the fact that E-E type knowledge is necessarily encoded within the decoder of the LLM backbone; consequently, leaving this component unmodified naturally prevents overfitting to the editing objective through the contamination of E-E knowledge. Nevertheless, FT-Vis still fails to achieve satisfactory performance on tasks such as OBP and NBG, and continues to exhibit deficiencies on basic metrics such as locality.

\textbf{Conclusion.}
Taken together, EC-Bench reveals that the apparent success of current MKE methods often conceals a inconsistent internal knowledge structure: (1) the original image-to-entity pathway $(i,e)$ remains active, (2) the new image-to-entity pathway $(i,e^*)$ is weak and difficult to be leveraged for complex reasoning, and (3) an unintended entity-level shortcut between $e$ and $e^*$ is introduced in the language space. When querying the model, it still perceives the image $i$ as the original entity $e$, and then exploits the shortcut $(e,e^*)$ to output the label of $e^*$, thereby creating the illusion of a successful edit.

% Overall, the combined EC-Bench profile shows that current MKE methods are closer to partial behavioral patching than principled multimodal knowledge replacement. These results motivate the mechanistic discussion in the next section.

% 这节需要两个实验的表：FT不同层数 折线图 + FT浅层 FT深层 MEND浅层深层 表格？

\section{Mitigating Entity Identity Confusion: A Preliminary Exploration}

The above analysis suggests that the lack of explicit distinction between I-E and E-E type knowledge in existing editing strategies likely leads models to incorrectly fit editing targets by forcibly altering E-E associations, rather than modifying the intended I-E binding relationships. We therefore argue that \textbf{a principled editing strategy should decouple these two types of knowledge, ensuring that editing interventions precisely target I-E binding representations while preserving the structural integrity of E-E associative knowledge}.

To address this issue, inspired by the observation that methods targeting visual modules exhibit significantly less severe EIC phenomenon, we hypothesize that controlling the location of the editing target module may serve as a minimalist yet effective mitigation strategy. In this section, we aim to conduct a preliminary exploratory analysis of EIC mitigation strategies, thereby providing methodological guidance for future research. We first introduce the theoretical foundations underlying the proposed mitigation strategy, then present empirical evidence of its effectiveness, and finally discuss the broader implications for future research directions.

\subsection{Background and Rationale: Knowledge Recall in LLMs}

\textbf{Two-Stage Knowledge Recall in LLMs.}  Recent interpretability research on both LLMs and LVLMs \citep{geva-etal-2021-transformer, geva-etal-2023-dissecting, venhoff2025laterecallexplainingtwohop} has outlined a common two-stage pipeline for knowledge recall. As individual tokens carry only partial, locally-scoped semantic content, attention modules in shallow layers first aggregate scattered token representations into a unified \emph{entity representation} that encodes the entity identity referred to by the input; mid-layer MLPs then inject relevant factual knowledge based on this representation, which is subsequently extracted in deeper layers for downstream reasoning \citep{meng2022locating, geva-etal-2023-dissecting, ye2025llmunlearningformindependent}. Specifically for LVLMs, visual tokens are first aggregated into a coherent entity representation---a process that corresponds precisely to the I-E binding most central to MKE, in the shallow layers of the LVLM, before any relational knowledge can be retrieved.

\textbf{Implications for MKE.}  This two-stage structure has direct implications for knowledge editing. Based on this, we posit that \textbf{if editing interventions are applied at layers \emph{before} the entity representation is fully consolidated, the edit is more likely to target the I-E binding pathway rather than disrupting downstream E-E relational knowledge decoding.} Conversely, editing deeper layers -- as most existing MKE methods do, likely perturbs relation decoding while leaving upstream binding intact, which is precisely the failure pattern we observe in EIC. We therefore propose that controlling the editing location may be a potentially effective strategy for multimodal knowledge editing.

\subsection{Mitigating EIC via Editing-Location Control}

To validate this hypothesis, we use FT to edit different layers of LLaVA-1.5 and examine the EIC performance of the resulting edited models. The specific results are shown in Figure~\ref{fig:FT-location}. We summarize our observations as follows:

\begin{wrapfigure}{r}{0.55\textwidth}
    \centering
    \vspace{-1.5em}
    \includegraphics[width=1\linewidth]{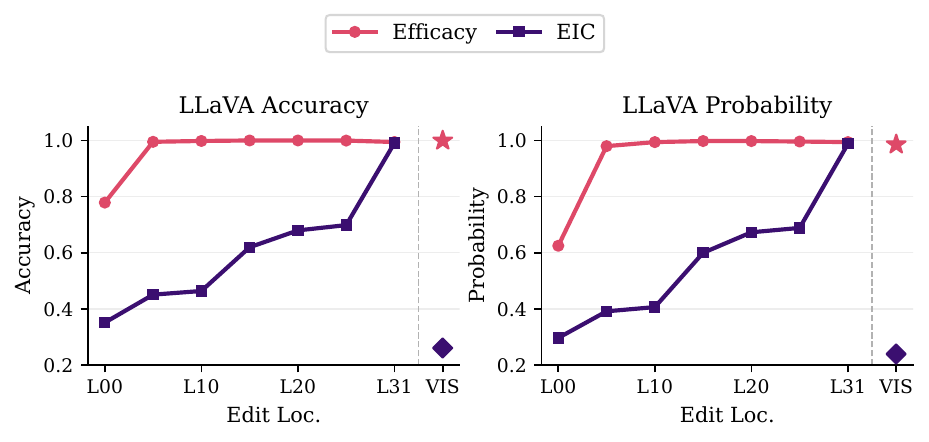}
    \caption{Results for FT on LLaVA with different editing locations.}
    \label{fig:FT-location}
    \vspace{-1em}
\end{wrapfigure}

\textbf{Obs1. Editing Shallow LLM Layers Reduces EIC.} As shown in Figure~\ref{fig:FT-location}, the model's EIC performance exhibits a strong correlation with the editing location: the severity of EIC increases monotonically as the edited layer moves deeper. We observe that editing shallow LLM layers does not produce severe EIC, yielding levels close to those of FT-Vis and the original model, suggesting that the shallow layers of the LLM backbone can still preserve textual entity identity. In contrast, at deeper layers, EIC reaches extremely high levels approaching 100\%, implying that the model may have completely overfit and lost its normal capacity for processing entity knowledge. To be noted, editing layer 0 results in a slight drop in edit success rate, which may be attributed to the inevitable negative perturbation that fine-tuning introduces to that layer's parameters; and layer 0 is particularly critical as it directly processes the input.

% Auto-generated from whitelist-backed results.
\begin{table*}[t]
\centering

\footnotesize
\setlength{\tabcolsep}{2.8pt}
\renewcommand{\arraystretch}{1.06}

\caption{Editing-location comparison for FT and MEND.}
\label{tab:edit-location}
\sf
\resizebox{\textwidth}{!}{%

\begin{tabular}{llcccccccc}
\toprule
Model & Method & Efficacy $\uparrow$ & T-Gen $\uparrow$ & I-Gen $\uparrow$ & T-Loc $\uparrow$ & I-Loc $\uparrow$ & EIC $\downarrow$ & OBP $\downarrow$ & NBG $\uparrow$ \\

\midrule

\multirow[c]{8}{*}{LLaVA} 
& base (unedited) & 32.4 & 35.8 & 32.1 & 100.0 & 100.0 & 26.1 & 92.2 & 40.4 \\
\cmidrule(lr){2-10}
& FT-Shallow & 99.5 & 92.9 & 97.1 & 80.5 & 38.7 & 45.1 & 68.8 & 53.4 \\ %l5
& FT-Deep & 99.4 & 99.1 & 99.4 & 59.3 & 24.3 & 99.1 & 72.3 & 33.7 \\
& FT-Vis & 99.9 & 98.1 & 96.4 & 100.0 & 18.8 & 26.1 & 51.2 & 54.6 \\
\cmidrule(lr){2-10}
& MEND-Shallow & 98.6 & 97.5 & 97.8 & 88.9 & 97.2 & 59.7 & 92.6 & 43.9 \\
& MEND-Deep & 99.0 & 98.7 & 98.9 & 81.4 & 86.8 & 97.2 & 92.1 & 38.8 \\
& MEND-Vis & 70.7 & 71.2 & 68.6 & 100.0 & 48.1 & 26.1 & 87.6 & 54.3 \\
    
\midrule
\midrule

\multirow[c]{8}{*}{MiniGPT-4} 
& base (unedited) & 29.4 & 32.1 & 29.1 & 100.0 & 100.0 & 31.8 & 73.0 & 38.1 \\
\cmidrule(lr){2-10}
& FT-Shallow & 82.2 & 73.7 & 76.4 & 90.8 & 60.6 & 42.3 & 76.2 & 46.3 \\ %l10
& FT-Deep & 100.0 & 99.5 & 99.9 & 83.2 & 47.1 & 69.7 & 51.0 & 38.9 \\
& FT-Vis & 100.0 & 99.5 & 100.0 & 100.0 & 19.8 & 31.8 & 65.5 & 39.5 \\
\cmidrule(lr){2-10}
& MEND-Shallow & 87.2 & 87.7 & 86.7 & 82.3 & 76.9 & 42.9 & 72.0 & 42.0 \\
& MEND-Deep & 99.4 & 99.1 & 99.3 & 82.6 & 85.5 & 94.2 & 76.9 & 37.6 \\
& MEND-Vis & 100.0 & 99.5 & 100.0 & 100.0 & 19.8 & 31.8 & 65.5 & 39.5 \\
\bottomrule
    
\end{tabular}}

\end{table*}

\textbf{Obs2. The Shape of the Curve Corroborates the Entity Representation Solidification Hypothesis.} Another aspect of the experiment lies in the shape of the curve, which provides supporting evidence for our theoretical framework. The severity of EIC does not increase linearly with editing depth: the curve remains relatively flat across the first few layers, its slope rises markedly upon entering the middle layers of the model, and becomes very steep in the deeper layers. We posit that this abrupt transition point likely corresponds to the layer at which entity representations solidify: before this point, edits primarily act on the image-to-entity (I-E) binding pathway; after this point, edits primarily disrupt downstream relation decoding, giving rise to the characteristic identity confusion of EIC. Interestingly, the layers implicated by EIC closely align with those identified in prior mechanistic interpretability work \citep{venhoff2025laterecallexplainingtwohop}, further corroborating the consistency between our EIC framework and established mechanistic understanding of entity representation formation in transformer models.

\textbf{Generalization to Other Methods.} A natural question is whether this finding is specific to FT or generalizes across editing paradigms. Since methods such as IKE and SERAC rely on external prompts and modules and do not involve editing specific layers, we select MEND -- a representative of another parameter-modification paradigm, and apply it to vision-side modules as well as shallow LLM layers. We observe the same mitigation effect: as shown in Table \ref{tab:edit-location}, MEND with edits confined to shallow or vision-side layers achieves a similarly significant reduction in EIC compared to the default deep-layer configuration. This suggests that shallow-layer editing is a generalizable principle and can serve as a design reference for parameter-modification-based knowledge editing methods. We also note that the improvements brought by shallow-layer editing on the OBP and NBG tasks remain less pronounced than on EIC itself, which further reflects that multi-hop reasoning in multimodal settings may be equally challenging as in text-only settings and warrants further exploration in future work.

\subsection{Discussion \& Implications for Future Research}

In conclusion, our analysis suggests that faithful  should distinguish between I-E binding, which must be updated, and E-E relational knowledge, which should remain intact. Meanwhile, OBP and NBG tasks remain equally challenging to resolve; given that multi-hop reasoning in the text-only modality is still an open problem, how to achieve truly faithful multimodal editing warrants further exploration in future work.

We frame our analysis of editing location as an exploration in this direction, and the robust reduction of EIC it yields indicates that editing location can still serve as a useful design principle for future MKE frameworks. More broadly, we believe that effective MKE requires further attention to diagnostic evaluations beyond surface efficacy, and calls for better mechanisms that localize edits to the appropriate representational stages.

\section{Related Work}

\textbf{Knowledge Editing in Large Language Models.} Knowledge editing aims to update model knowledge precisely and efficiently while preserving unrelated knowledge intact \citep{zhang2024comprehensivestudyknowledgeediting,wang2023easyedit}. Knowledge editing methods can be broadly categorized into two types \citep{zhang2024comprehensivestudyknowledgeediting,zhang-etal-2025-kele,zhang2025uncovering,zhou2026uncoveringcontextrelianceunstructured}: \textit{parameter-modifying} methods directly modify internal weights to enforce the injection of target factsl for example, FT directly fine-tunes model parameters; KE \citep{de-cao-etal-2021-editing} and MEND \citep{mitchell2022fast} train a hypernetwork to generate parameter updates; ROME \citep{meng2022locating}, MEMIT \citep{meng2023massediting}, and GLAME \citep{zhang-etal-2024-knowledge-graph} first locate knowledge storage positions before performing targeted updates. \textit{Parameter-preserving} methods rewrite model behavior through retrieval or external memory; IKE \citep{zheng-etal-2023-edit} alters model outputs via in-context learning, while memory-based methods such as SERAC \citep{mitchell2022memory} modify model behavior through an additional memory module.

\textbf{Multimodal Knowledge Editing.} Recent research on multimodal editing has extended the knowledge editing paradigm to LVLMs, migrating a series of editing methods to LVLMs \citep{NEURIPS2024VLKEB,NEURIPS2024UNIKE,Zeng_2025_ICCV} and producing a range of benchmark works, such as the representative datasets MMEdit \citep{cheng2023edit}, MIKE \citep{li-etal-2024-mike}, VLKEB \citep{NEURIPS2024VLKEB}, and MC-MKE \citep{zhang-etal-2025-mc}, forming an evaluation framework centered on efficacy, generalization, and locality. However, current evaluations of MKE remain dominated by surface-level efficacy on simple questions, with insufficient analysis of the specific behavioral patterns of edited models. This allows many methods with underlying issues to still achieve favorable results. Our work provides a valuable complement to this line of research and reveals that high efficacy scores may conceal severe internal knowledge inconsistency.

% \textbf{Knowledge Mechanisms in LLMs.} A growing body of analysis suggests that knowledge in LVLMs is distributed across heterogeneous modules and layers. Early visual alignment modules are crucial for mapping visual evidence to semantic entities, while deeper language layers are responsible for relation-level reasoning and answer generation. This hierarchical perspective aligns closely with the findings of this paper: editing behavior is strongly dependent on the intervention location, and edits targeting a single layer or module often fail to achieve coherent binding replacement. Accordingly, this paper advocates for binding-aware and structure-aware editing strategies that accommodate the intrinsic characteristics of representational heterogeneity.

\section{Conclusion}

In this work, we identified and characterized Entity Identity Confusion (EIC), a systemic yet previously overlooked failure mode in multimodal knowledge editing that existing benchmarks largely fail to detect. We demonstrated that EIC stems from the failure of current MKE methods to distinguish between I-E and E-E knowledge, leading models to overfit E-E associations as a shortcut rather than the underlying I-E binding. To rigorously diagnose this phenomenon, we introduced EC-Bench, a benchmark featuring challenging tasks that expose EIC where standard evaluations cannot.

Building on our mechanistic analysis, we identified constraining edits to early-stage representations as a promising mitigation direction, and discussed the principled desiderata that a faithful MKE method should satisfy. We hope the problem formulation, benchmark, and insights presented here provide a useful foundation for future research toward more faithful and robust multimodal knowledge editing.

% \begin{ack}
% Use unnumbered first level headings for the acknowledgments. All acknowledgments

% Do {\bf not} include this section in the anonymized submission, only in the final paper. You can use the \texttt{ack} environment provided in the style file to automatically hide this section in the anonymized submission.
% \end{ack}

\bibliography{main}

@article{meng2022locating,
  title={Locating and editing factual associations in GPT},
  author={Meng, Kevin and Bau, David and Andonian, Alex and Belinkov, Yonatan},
  journal={Advances in Neural Information Processing Systems},
  volume={35},
  pages={17359--17372},
  year={2022}
}

@misc{zhao2025surveylargelanguagemodels,
      title={A Survey of Large Language Models}, 
      author={Wayne Xin Zhao and Kun Zhou and Junyi Li and Tianyi Tang and Xiaolei Wang and Yupeng Hou and Yingqian Min and Beichen Zhang and others},
      year={2025},
      eprint={2303.18223},
      archivePrefix={arXiv},
      primaryClass={cs.CL},
      url={https://arxiv.org/abs/2303.18223}, 
}

@inproceedings{geva-etal-2023-dissecting,
    title = "Dissecting Recall of Factual Associations in Auto-Regressive Language Models",
    author = "Geva, Mor  and
      Bastings, Jasmijn  and
      Filippova, Katja  and
      Globerson, Amir",
    editor = "Bouamor, Houda  and
      Pino, Juan  and
      Bali, Kalika",
    booktitle = "Proceedings of the 2023 Conference on Empirical Methods in Natural Language Processing",
    month = dec,
    year = "2023",
    address = "Singapore",
    publisher = "Association for Computational Linguistics",
    url = "https://aclanthology.org/2023.emnlp-main.751/",
    doi = "10.18653/v1/2023.emnlp-main.751",
    pages = "12216--12235"
}

@inproceedings{
meng2023massediting,
title={Mass-Editing Memory in a Transformer},
author={Kevin Meng and Arnab Sen Sharma and Alex J Andonian and Yonatan Belinkov and David Bau},
booktitle={The Eleventh International Conference on Learning Representations },
year={2023}
}

@misc{zhang2024comprehensivestudyknowledgeediting,
      title={A Comprehensive Study of Knowledge Editing for Large Language Models}, 
      author={Ningyu Zhang and Yunzhi Yao and Bozhong Tian and Peng Wang and Shumin Deng and Mengru Wang and Zekun Xi and Shengyu Mao and Jintian Zhang and Yuansheng Ni and Siyuan Cheng and Ziwen Xu and Xin Xu and Jia-Chen Gu and Yong Jiang and Pengjun Xie and Fei Huang and Lei Liang and Zhiqiang Zhang and Xiaowei Zhu and Jun Zhou and Huajun Chen},
      year={2024},
      eprint={2401.01286},
      archivePrefix={arXiv},
      primaryClass={cs.CL},
      url={https://arxiv.org/abs/2401.01286}, 
}

@inproceedings{geva-etal-2021-transformer,
    title = "Transformer Feed-Forward Layers Are Key-Value Memories",
    author = "Geva, Mor  and
      Schuster, Roei  and
      Berant, Jonathan  and
      Levy, Omer",
    editor = "Moens, Marie-Francine  and
      Huang, Xuanjing  and
      Specia, Lucia  and
      Yih, Scott Wen-tau",
    booktitle = "Proceedings of the 2021 Conference on Empirical Methods in Natural Language Processing",
    month = nov,
    year = "2021",
    address = "Online and Punta Cana, Dominican Republic",
    publisher = "Association for Computational Linguistics",
    url = "https://aclanthology.org/2021.emnlp-main.446/",
    doi = "10.18653/v1/2021.emnlp-main.446",
    pages = "5484--5495"
}

@inproceedings{zhang-etal-2024-knowledge-graph,
    title = "Knowledge Graph Enhanced Large Language Model Editing",
    author = "Zhang, Mengqi  and
      Ye, Xiaotian  and
      Liu, Qiang  and
      Ren, Pengjie  and
      Wu, Shu  and
      Chen, Zhumin",
    editor = "Al-Onaizan, Yaser  and
      Bansal, Mohit  and
      Chen, Yun-Nung",
    booktitle = "Proceedings of the 2024 Conference on Empirical Methods in Natural Language Processing",
    month = nov,
    year = "2024",
    address = "Miami, Florida, USA",
    publisher = "Association for Computational Linguistics",
    url = "https://aclanthology.org/2024.emnlp-main.1261/",
    doi = "10.18653/v1/2024.emnlp-main.1261",
    pages = "22647--22662"
}

@inproceedings{
    zhang2025uncovering,
    title={Uncovering Overfitting in Large Language Model Editing},
    author={Mengqi Zhang and Xiaotian Ye and Qiang Liu and Shu Wu and Pengjie Ren and Zhumin Chen},
    booktitle={The Thirteenth International Conference on Learning Representations},
    year={2025},
    url={https://openreview.net/forum?id=t8qcGXaepr}
}

@misc{ye2025llmunlearningformindependent,
      title={LLM Unlearning Should Be Form-Independent}, 
      author={Xiaotian Ye and Mengqi Zhang and Shu Wu},
      year={2025},
      eprint={2506.07795},
      archivePrefix={arXiv},
      primaryClass={cs.CL},
      url={https://arxiv.org/abs/2506.07795}, 
}

@inproceedings{zhang-etal-2025-kele,
    title = "{KELE}: Residual Knowledge Erasure for Enhanced Multi-hop Reasoning in Knowledge Editing",
    author = "Zhang, Mengqi  and
      Fang, Bowen  and
      Liu, Qiang  and
      Ye, Xiaotian  and
      Wu, Shu  and
      Ren, Pengjie  and
      Chen, Zhumin  and
      Wang, Liang",
    editor = "Christodoulopoulos, Christos  and
      Chakraborty, Tanmoy  and
      Rose, Carolyn  and
      Peng, Violet",
    booktitle = "Findings of the Association for Computational Linguistics: EMNLP 2025",
    month = nov,
    year = "2025",
    address = "Suzhou, China",
    publisher = "Association for Computational Linguistics",
    url = "https://aclanthology.org/2025.findings-emnlp.1334/",
    doi = "10.18653/v1/2025.findings-emnlp.1334",
    pages = "24537--24552",
    ISBN = "979-8-89176-335-7"
}

@misc{zhang2026spectralcharacterizationmitigationsequential,
      title={Spectral Characterization and Mitigation of Sequential Knowledge Editing Collapse}, 
      author={Chi Zhang and Mengqi Zhang and Xiaotian Ye and Runxi Cheng and Zisheng Zhou and Ying Zhou and Pengjie Ren and Zhumin Chen},
      year={2026},
      eprint={2601.11042},
      archivePrefix={arXiv},
      primaryClass={cs.CL},
      url={https://arxiv.org/abs/2601.11042}, 
}

@inproceedings{liu2019mmkg,
  title={MMKG: multi-modal knowledge graphs},
  author={Liu, Ye and Li, Hui and Garcia-Duran, Alberto and Niepert, 
  Mathias and Onoro-Rubio, Daniel and Rosenblum, David S},
  booktitle={European Semantic Web Conference},
  pages={459--474},
  year={2019},
  organization={Springer}
}

@inproceedings{mitchell2022fast,
    title={Fast Model Editing at Scale},
    author={Eric Mitchell and Charles Lin and Antoine Bosselut and Chelsea Finn and Christopher D Manning},
    booktitle={International Conference on Learning Representations},
    year={2022},
    url={https://openreview.net/pdf?id=0DcZxeWfOPt}
}

@inproceedings{mitchell2022memory,
    title={Memory-Based Model Editing at Scale},
    author={Mitchell, Eric and Lin, Charles and Bosselut, Antoine and Finn, Chelsea and Manning, Christopher D.},
    booktitle={International Conference on Machine Learning},
    url={https://arxiv.org/pdf/2206.06520.pdf},
    year={2022},
}

@inproceedings{zheng-etal-2023-edit,
    title = "Can We Edit Factual Knowledge by In-Context Learning?",
    author = "Zheng, Ce  and
      Li, Lei  and
      Dong, Qingxiu  and
      Fan, Yuxuan  and
      Wu, Zhiyong  and
      Xu, Jingjing  and
      Chang, Baobao",
    editor = "Bouamor, Houda  and
      Pino, Juan  and
      Bali, Kalika",
    booktitle = "Proceedings of the 2023 Conference on Empirical Methods in Natural Language Processing",
    month = dec,
    year = "2023",
    address = "Singapore",
    publisher = "Association for Computational Linguistics",
    url = "https://aclanthology.org/2023.emnlp-main.296",
    doi = "10.18653/v1/2023.emnlp-main.296",
    pages = "4862--4876",
}

@article{cheng2023edit,
  title={Can We Edit Multimodal Large Language Models?}, 
  author={Cheng, Siyuan and Tian, Bozhong and Liu, Qingbin and Chen, Xi and Wang, Yongheng and Chen, Huajun and Zhang, Ningyu},
  journal={arXiv preprint arXiv:2310.08475},
  year={2023}
}

@inproceedings{de-cao-etal-2021-editing,
    title = "Editing Factual Knowledge in Language Models",
    author = "De Cao, Nicola  and
      Aziz, Wilker  and
      Titov, Ivan",
    editor = "Moens, Marie-Francine  and
      Huang, Xuanjing  and
      Specia, Lucia  and
      Yih, Scott Wen-tau",
    booktitle = "Proceedings of the 2021 Conference on Empirical Methods in Natural Language Processing",
    month = nov,
    year = "2021",
    address = "Online and Punta Cana, Dominican Republic",
    publisher = "Association for Computational Linguistics",
    url = "https://aclanthology.org/2021.emnlp-main.522",
    doi = "10.18653/v1/2021.emnlp-main.522",
    pages = "6491--6506",
}

@misc{ye2023mplugowl2,
      title={mPLUG-Owl2: Revolutionizing Multi-modal Large Language Model with Modality Collaboration}, 
      author={Qinghao Ye and Haiyang Xu and Jiabo Ye and Ming Yan and Anwen Hu and Haowei Liu and Qi Qian and Ji Zhang and Fei Huang and Jingren Zhou},
      year={2023},
      eprint={2311.04257},
      archivePrefix={arXiv},
      primaryClass={cs.CL}
}

@article{liu2023llava,
  title={Visual Instruction Tuning},
  author={Liu, Haotian and Li, Chunyuan and Wu, Qingyang and Lee, Yong Jae},
  journal={arXiv preprint arXiv:2304.08485},
  year={2023},
  url={https://arxiv.org/abs/2304.08485}
}

@inproceedings{cheng-etal-2023-edit,
    title = "Can We Edit Multimodal Large Language Models?",
    author = "Cheng, Siyuan  and
      Tian, Bozhong  and
      Liu, Qingbin  and
      Chen, Xi  and
      Wang, Yongheng  and
      Chen, Huajun  and
      Zhang, Ningyu",
    editor = "Bouamor, Houda  and
      Pino, Juan  and
      Bali, Kalika",
    booktitle = "Proceedings of the 2023 Conference on Empirical Methods in Natural Language Processing",
    month = dec,
    year = "2023",
    address = "Singapore",
    publisher = "Association for Computational Linguistics",
    url = "https://aclanthology.org/2023.emnlp-main.856",
    doi = "10.18653/v1/2023.emnlp-main.856",
    pages = "13877--13888",
}

@article{Qwen-VL,
  title={Qwen-VL: A Versatile Vision-Language Model for Understanding, Localization, Text Reading, and Beyond},
  author={Bai, Jinze and Bai, Shuai and Yang, Shusheng and Wang, Shijie and Tan, Sinan and Wang, Peng and Lin, Junyang and Zhou, Chang and Zhou, Jingren},
  journal={arXiv preprint arXiv:2308.12966},
  year={2023}
}

@article{wang2023easyedit,
  title={Easyedit: An easy-to-use knowledge editing framework for large language models},
  author={Wang, Peng and Zhang, Ningyu and Xie, Xin and Yao, Yunzhi and Tian, Bozhong and Wang, Mengru and Xi, Zekun and Cheng, Siyuan and Liu, Kangwei and Zheng, Guozhou and others},
  journal={arXiv preprint arXiv:2308.07269},
  year={2023}
}

@inproceedings{li2023blip2,
      title={{BLIP-2:} Bootstrapping Language-Image Pre-training with Frozen Image Encoders and Large Language Models}, 
      author={Junnan Li and Dongxu Li and Silvio Savarese and Steven Hoi},
      year={2023},
      booktitle={ICML},
}

@article{zhu2023minigpt,
  title={MiniGPT-4: Enhancing Vision-Language Understanding with Advanced Large Language Models},
  author={Zhu, Deyao and Chen, Jun and Shen, Xiaoqian and Li, Xiang and Elhoseiny, Mohamed},
  journal={arXiv preprint arXiv:2304.10592},
  year={2023}
}

@inproceedings{NEURIPS2024VLKEB,
 author = {Huang, Han and Zhong, Haitian and Yu, Tao and Liu, Qiang and Wu, Shu and Wang, Liang and Tan, Tieniu},
 booktitle = {Advances in Neural Information Processing Systems},
 doi = {10.52202/079017-0294},
 editor = {A. Globerson and L. Mackey and D. Belgrave and A. Fan and U. Paquet and J. Tomczak and C. Zhang},
 pages = {9257--9280},
 publisher = {Curran Associates, Inc.},
 title = {VLKEB: A Large Vision-Language Model Knowledge Editing Benchmark},
 url = {https://proceedings.neurips.cc/paper_files/paper/2024/file/1198b53fa686831d5f0c0860d7ec4f34-Paper-Datasets_and_Benchmarks_Track.pdf},
 volume = {37},
 year = {2024}
}

@misc{venhoff2025laterecallexplainingtwohop,
      title={Too Late to Recall: Explaining the Two-Hop Problem in Multimodal Knowledge Retrieval}, 
      author={Constantin Venhoff and Ashkan Khakzar and Sonia Joseph and Philip Torr and Neel Nanda},
      year={2025},
      eprint={2512.03276},
      archivePrefix={arXiv},
      primaryClass={cs.LG},
      url={https://arxiv.org/abs/2512.03276}, 
}

@inproceedings{NEURIPS2024UNIKE,
 author = {Pan, Kaihang and Fan, Zhaoyu and Li, Juncheng and Yu, Qifan and Fei, Hao and Tang, Siliang and Hong, Richang and Zhang, Hanwang and Sun, Qianru},
 booktitle = {Advances in Neural Information Processing Systems},
 doi = {10.52202/079017-3500},
 editor = {A. Globerson and L. Mackey and D. Belgrave and A. Fan and U. Paquet and J. Tomczak and C. Zhang},
 pages = {110290--110314},
 publisher = {Curran Associates, Inc.},
 title = {Towards Unified Multimodal Editing with Enhanced Knowledge Collaboration},
 url = {https://proceedings.neurips.cc/paper_files/paper/2024/file/c705ba25f183b875c9359ef83fa262e8-Paper-Conference.pdf},
 volume = {37},
 year = {2024}
}

@InProceedings{Zeng_2025_ICCV,
    author    = {Zeng, Zhen and Gu, Leijiang and Yang, Xun and Duan, Zhangling and Shi, Zenglin and Wang, Meng},
    title     = {Visual-Oriented Fine-Grained Knowledge Editing for MultiModal Large Language Models},
    booktitle = {Proceedings of the IEEE/CVF International Conference on Computer Vision (ICCV)},
    month     = {October},
    year      = {2025},
    pages     = {2491-2500}
}

@inproceedings{li-etal-2024-mike,
    title = "{MIKE}: A New Benchmark for Fine-grained Multimodal Entity Knowledge Editing",
    author = "Li, Jiaqi  and
      Du, Miaozeng  and
      Zhang, Chuanyi  and
      Chen, Yongrui  and
      Hu, Nan  and
      Qi, Guilin  and
      Jiang, Haiyun  and
      Cheng, Siyuan  and
      Tian, Bozhong",
    editor = "Ku, Lun-Wei  and
      Martins, Andre  and
      Srikumar, Vivek",
    booktitle = "Findings of the Association for Computational Linguistics: ACL 2024",
    month = aug,
    year = "2024",
    address = "Bangkok, Thailand",
    publisher = "Association for Computational Linguistics",
    url = "https://aclanthology.org/2024.findings-acl.298/",
    doi = "10.18653/v1/2024.findings-acl.298",
    pages = "5018--5029"
}

@inproceedings{zhang-etal-2025-mc,
    title = "{MC}-{MKE}: A Fine-Grained Multimodal Knowledge Editing Benchmark Emphasizing Modality Consistency",
    author = "Zhang, Junzhe  and
      Zhang, Huixuan  and
      Yin, Xunjian  and
      Huang, Baizhou  and
      Zhang, Xu  and
      Hu, Xinyu  and
      Wan, Xiaojun",
    editor = "Che, Wanxiang  and
      Nabende, Joyce  and
      Shutova, Ekaterina  and
      Pilehvar, Mohammad Taher",
    booktitle = "Findings of the Association for Computational Linguistics: ACL 2025",
    month = jul,
    year = "2025",
    address = "Vienna, Austria",
    publisher = "Association for Computational Linguistics",
    url = "https://aclanthology.org/2025.findings-acl.896/",
    doi = "10.18653/v1/2025.findings-acl.896",
    pages = "17430--17445",
    ISBN = "979-8-89176-256-5"
}

@misc{zhou2026uncoveringcontextrelianceunstructured,
      title={Uncovering Context Reliance in Unstructured Knowledge Editing}, 
      author={Zisheng Zhou and Mengqi Zhang and Shiguang Wu and Xiaotian Ye and Chi Zhang and Zhumin Chen and Pengjie Ren},
      year={2026},
      eprint={2602.19043},
      archivePrefix={arXiv},
      primaryClass={cs.CL},
      url={https://arxiv.org/abs/2602.19043}, 
}
\bibliographystyle{plainnat}

%%%%%%%%%%%%%%%%%%%%%%%%%%%%%%%%%%%%%%%%%%%%%%%%%%%%%%%%%%%%

\newpage
\appendix

\section{Limitations}
\label{apd:limitation}

EC-Bench follows the classical MKE setting and primarily focuses on the binding relationship between images and entities. Accordingly, the scope of this paper is largely confined to analyzing the behavioral patterns of edited models under this type of multimodal editing. However, real-world multimodal scenarios may involve more complex tasks, including a richer variety of entity categories, image types, and knowledge beyond entity-level understanding, such as knowledge of image style. Future work may extend EC-Bench to broader real-world image distributions, more diverse entity types, and explore knowledge categories beyond entity knowledge.

\section{Impact Statement}
\label{apd:statement}

This paper presents work whose goal is to advance the field of multimodal knowledge editing for large vision-language models. By identifying and formalizing Entity Identity Confusion (EIC) as a systemic failure mode, introducing EC-Bench as a diagnostic benchmark, and proposing editing-location control as a principled mitigation strategy, our work improves the transparency and reliability of multimodal knowledge editing. These contributions help ensure that knowledge updates in deployed LVLMs are faithful and consistent, rather than producing superficially correct yet internally corrupted behavior.

While multimodal knowledge editing has broad positive applications, including correcting outdated information, enforcing safety policies, and enabling continual model maintenance, we acknowledge potential ethical considerations. The ability to alter a model's internal knowledge bindings could be misused to inject biased or misleading associations between visual content and entity identities. We encourage future work to develop safeguards against such misuse and to ensure that multimodal knowledge editing techniques are deployed in alignment with ethical AI principles.

\section{Details on EC-Bench Benchmark}
\label{apd:ecbench}

EC-Bench is a benchmark designed for evaluate the effectiveness of multimodal knowledge editing methods for LVLMs. Its goal is to detect EIC and evaluate how the bindings between images and the original/new entities evolve after editing, thereby allows us to analyze the characteristics of EIC and analyze its underlying mechanisms. 

To be noted, part of the EC-Bench dataset are sourced from existing open-source MKE benchmarks. We have specifically restructured and extended the VLKEB \citep{NEURIPS2024VLKEB} dataset, constructing new data and added novel tasks, while building upon its original baselines and hyperparameter settings to enable a more comprehensive and in-depth evaluation of the EIC issue we investigate.

\subsection{Dataset Composition}

At the benchmark level, the base data unit is a counterfactual image--entity edit tuple
\begin{equation}
(i,e)\rightarrow(i,e^*),
\end{equation}
where $i$ is an image originally associated with entity $e$, and the edited model is expected to recognize the same image under the target entity $e^*$.

For each edit tuple, EC-Bench instantiates a set of fundamental tasks, including the original edit query, a text-side rephrasing, an image-side rephrasing, and locality examples. In addition to these standard evaluation dimensions, EC-Bench includes three Binding Diagnostic Tasks. EIC is the central diagnostic in EC-Bench and measures whether the edit creates language-side entity confusion between $e$ and $e^*$; OBP measures whether the original image--entity binding remains active after editing; and NBG evaluates whether the edited image supports target-side relational reasoning about $e^*$. The scale of the current evaluation split used in our experiments is summarized in Table~\ref{tab:appendix-ecbench-scale}.

\begin{table}[h!]
\centering
% \footnotesize
\setlength{\tabcolsep}{6pt}
\renewcommand{\arraystretch}{1.08}
\caption{Scale summary of the current EC-Bench evaluation split used in our experiments.}
\label{tab:appendix-ecbench-scale}
\sf
\begin{tabular}{@{}lc@{}}
\toprule
Item & Count \\
\midrule
Edit cases & 1545 \\
Original edit-query images & 1545 \\
Image-side generalization images & 1455 \\
Locality images & 1193 \\
NBG questions & 2464 \\
EIC questions & 1545 \\
OBP questions & 1545 \\
\bottomrule
\end{tabular}
\end{table}

\subsection{Dataset Construction Details}

The edit tuples and fundamental task data in EC-Bench are sourced from the established benchmark VLKEB \citep{NEURIPS2024VLKEB}, specifically leveraging a subset of their test splits. We further constructing new data and added novel binding diagnostic tasks for assess post-edit multi-modal binding behavior.

As a concrete running example, consider an edit case in which the original image is associated with \emph{Jack Webb} and the target entity is \emph{Joseph Cotten}. The corresponding edit tuple is
\begin{equation}
(i,\text{Jack Webb}) \rightarrow (i,\text{Joseph Cotten}).
\end{equation}
The examples below instantiate EIC, OBP, and NBG around this same edit tuple before presenting additional real evaluation cases.

\paragraph{EIC construction.}
The goal of EIC is to test whether the edited model incorrectly connects the original entity $e$ to the target entity $e^*$ on the language side. The construction starts from the image-conditioned entity-identification question already present in Efficacy Task. Let this source question be denoted by $q_{\mathrm{img}}(i)$, whose answer in the original data is the entity $e$. We convert it into a text-only question by replacing the image referent in the question with the exact entity name, while keeping the remaining semantics unchanged. The resulting EIC question is a text-only question about $e$:
\begin{equation}
q_{\mathrm{EIC}}(e)=\mathcal{R}(e, q_{\mathrm{img}}),
\end{equation}
where $\mathcal{R}$ denotes the rewrite operation performed by an external LLM. In the Jack Webb example, the source image-conditioned question is
\begin{equation}
q_{\mathrm{img}}(i)=\text{``Who is the actor featured in this image?''}
\end{equation}
and the rewritten EIC question becomes
\begin{equation}
q_{\mathrm{EIC}}(e)=\text{``Who is the actor Jack Webb?''}
\end{equation}
with target answer \emph{Joseph Cotten}. If the edited model answers the question about Jack Webb with Joseph Cotten, the edit has introduced entity confusion.

We use \texttt{DeepSeek-Chat} as the rewriting model. The prompt is designed to preserve the semantic structure of the original image-conditioned question while removing image dependence and directly inserting the original entity name. The full prompt used in construction is shown below.

\begin{tcolorbox}[
  breakable,
  colback=blue!3,
  colframe=blue!45!black,
  title={Prompt Block A1: EIC question rewriting},
  fonttitle=\bfseries
]
\small
\textbf{Model.} \texttt{DeepSeek-Chat}

\textbf{System prompt.}

You are a powerful question-and-answer generator. The user provides a question about a specific entity (person, location, sign, scene, poster, logo, sight, occupation, etc.) in an image, where the answer to the question is the entity itself. Your task is to rewrite the question by replacing any reference to ``the image'' with the exact entity provided, while keeping the rest of the question unchanged and ensuring linguistic and logical coherence.

Output format: Directly output the rewritten question only. Do not include any additional information.

Examples:

User Input: Entity: Denton, Texas $\mid$ Question: What city of Texas is depicted in the image?

Output: What city of Texas is Denton, Texas?

User Input: Entity: Lily Tomlin $\mid$ Question: Who is the actor featured in this image?

Output: Who is the actor Lily Tomlin?

User Input: Entity: Carl Stalling $\mid$ Question: What is the full name of the musician depicted in the image?

Output: What is the full name of the musician Carl Stalling?

User Input: Entity: A.C. Milan $\mid$ Question: Which football club is represented in the image?

Output: Which football club is A.C. Milan?

\textbf{User template.}

Entity: \{entity\} $\mid$ Question: \{question\}
\end{tcolorbox}

\paragraph{OBP construction.}
The goal of OBP is to test whether the old I-E binding $(i,e)$ still survives after editing. To make this test discriminative, we construct a relation-controlled A/B question from an entity pair in which the original entity and the target entity correspond to different answers under the same relation:
\begin{equation}
(e,r,o), \qquad (e^*,r,o^*), \qquad o\neq o^*.
\end{equation}
This condition ensures that the question can distinguish whether the model is still following the original image-entity pathway or has shifted away from it. In this construction, the option associated with the original entity is always assigned to \texttt{A}, and the option associated with the target entity is assigned to \texttt{B}. The OBP question is then obtained by instantiating the resulting template with an image referent phrase, yielding an image-conditioned multiple-choice question whose old-binding answer remains \texttt{A} if the original binding is still active.

In the Jack Webb example, the generated relation-controlled question tests a role associated with the two entities. The resulting OBP question is:
\begin{quote}
\small
the man in this image is best known for playing the role of\\
\texttt{A. joe friday}\\
\texttt{B. holly martins}\\
Answer with one letter only (A or B):
\end{quote}
The old-binding answer is \texttt{A}. If the edited model still prefers \texttt{A} after editing, the original image--entity binding remains active.

We again use \texttt{DeepSeek-Chat} to construct these questions. The model receives the original entity and the target entity and is instructed to produce a question template, one answer option for the original entity, one answer option for the target entity, and a strong one-letter answer cue. The prompt explicitly constrains the generated question to be simple, factual, short, and non-ambiguous. The full construction prompt is shown below.

\begin{tcolorbox}[
  breakable,
  colback=green!3,
  colframe=green!45!black,
  title={Prompt Block A2: OBP question generation},
  fonttitle=\bfseries
]
\small
\textbf{Model.} \texttt{DeepSeek-Chat}

\textbf{Generation setting.} Temperature = 0.25

\textbf{System prompt.}

You are a precision multiple-choice question designer for multimodal knowledge-editing datasets. Given a pred entity and an alt entity, design a very simple question whose answer is a single letter: A or B.

Goals:
\begin{enumerate}[leftmargin=1.5em,itemsep=0.15em]
\item Choose a very simple property, relation, or description of the pred entity that makes the correct option highly predictable.
\item The question must strongly guide the model to output exactly one letter: A or B.
\item The question must end with a strong answer cue that explicitly requests one letter only.
\item Option A must be the correct completion for the pred entity.
\item Option B must be the correct completion for the alt entity.
\item Keep both options short, factual, and easy to read.
\item Use the exact four labels below.
\item Output exactly four lines:
\begin{itemize}[leftmargin=1.5em,itemsep=0.1em]
\item \texttt{question\_template: <sentence with \_\_SUBJECT\_\_ placeholder and no options>}
\item \texttt{answer\_a: <pred answer option>}
\item \texttt{answer\_b: <alt answer option>}
\item \texttt{question\_suffix: <strong final cue, for example ``Answer with one letter only (A or B):''>}
\end{itemize}
\end{enumerate}

Rules:
\begin{enumerate}[leftmargin=1.5em,itemsep=0.15em]
\item Do not use yes/no wording.
\item Do not output explanations.
\item Do not make the question too hard or too specific.
\item Do not use punctuation that makes the answer ambiguous.
\item Keep option text lowercase if possible.
\item Do not include the options inside the template line.
\end{enumerate}

Example 1:

Input:

pred entity: Monte Carlo

alt entity: Marrakesh

Output:

\texttt{question\_template: \_\_SUBJECT\_\_ is a district of the principality of}

\texttt{answer\_a: monaco}

\texttt{answer\_b: morocco}

\texttt{question\_suffix: Answer with one letter only (A or B):}

Example 2:

Input:

pred entity: Shine (film)

alt entity: Please Give

Output:

\texttt{question\_template: \_\_SUBJECT\_\_ is directed by}

\texttt{answer\_a: scott hicks}

\texttt{answer\_b: kirsten johnson}

\texttt{question\_suffix: Answer with one letter only (A or B):}

\textbf{User template.}

\texttt{pred entity: \{pred\_entity\}}

\texttt{alt entity: \{alt\_entity\}}
\end{tcolorbox}

The output of this prompt is a template with a subject placeholder \texttt{\_\_SUBJECT\_\_}, together with two short answer options. The OBP question is constructed by replacing the subject placeholder with an image referent phrase such as ``the man in this image'' or ``the film in this image''.

\paragraph{NBG construction.}
The goal of NBG is to test whether the new binding $(i,e^*)$ supports factual reasoning beyond the edited prompt. In practice, we use the portability tasks from the original VLKEB dataset as the data source for this task after filtering and simple processing, as both tasks involve multi-hop reasoning and essentially probe the same content. Each NBG example is an image-conditioned open-ended question whose answer is a factual attribute or relation associated with $e^*$.

\subsection{Examples of Dataset Entries}

The following examples are drawn from the EC-Bench Dataset.

\begin{tcolorbox}[breakable,colback=gray!2,colframe=gray!50,title={Example 1}]
\footnotesize
\begin{tabularx}{\linewidth}{>{\raggedright\arraybackslash\bfseries}p{0.27\linewidth}X}
Edit query & Who is the actor featured in this image? \\
Text generalization query & Which actor is shown in this picture? \\
Original entity & Jack Webb \\
Target entity & Joseph Cotten \\
EIC question & Who is the actor Jack Webb? \\
EIC target answer & Joseph Cotten \\
Image locality question & What educational institution is represented in the image? \\
Image locality answer & University of Florida \\
NBG question & \textbf{1-hop}: Where did the person associated with the individual in the picture pass away? \\
NBG target answer & Los Angeles \\
OBP question & \begin{minipage}[t]{\linewidth}\raggedright the man in this image is best known for playing the role of\\
A. joe friday\\
B. holly martins\\
Answer with one letter only (A or B):\end{minipage} \\
OBP old-binding answer & A \\
\end{tabularx}
\end{tcolorbox}

\begin{tcolorbox}[breakable,colback=gray!2,colframe=gray!50,title={Example 2}]
\footnotesize
\begin{tabularx}{\linewidth}{>{\raggedright\arraybackslash\bfseries}p{0.27\linewidth}X}
Edit query & What is the title of the movie in this scene? \\
Text generalization query & Which film does this image represent? \\
Original entity & The Full Monty \\
Target entity & Hustle \& Flow \\
EIC question & What is the title of the movie The Full Monty? \\
EIC target answer & Hustle \& Flow \\
Image locality question & What is the title of the film represented in the image? \\
Image locality answer & The Ladykillers (2004 film) \\
NBG question & \textbf{1-hop}: Which company distributed the film represented in the picture? \\
NBG target answer & Paramount Vantage \\
OBP question & \begin{minipage}[t]{\linewidth}\raggedright the film in this image is a film about a group of\\
A. unemployed steelworkers\\
B. aspiring rappers\\
Answer with one letter only (A or B):\end{minipage} \\
OBP old-binding answer & A \\
\end{tabularx}
\end{tcolorbox}

\begin{tcolorbox}[breakable,colback=gray!2,colframe=gray!50,title={Example 3}]
\footnotesize
\begin{tabularx}{\linewidth}{>{\raggedright\arraybackslash\bfseries}p{0.27\linewidth}X}
Edit query & What Irish county is depicted in the image? \\
Text generalization query & Which Irish county is seen in the picture? \\
Original entity & County Tipperary \\
Target entity & County Waterford \\
EIC question & What Irish county is County Tipperary? \\
EIC target answer & County Waterford \\
Image locality question & What city in Ireland is depicted in the image? \\
Image locality answer & Waterford \\
NBG question & \textbf{1-hop}: What types of political leadership can be found in the area neighboring the location shown in the picture? \\
NBG target answer & Fine Gael, Independent politician, Labour Party (Ireland), Social democracy \\
OBP question & \begin{minipage}[t]{\linewidth}\raggedright the place in this image is located in the province of\\
A. munster\\
B. leinster\\
Answer with one letter only (A or B):\end{minipage} \\
OBP old-binding answer & A \\
\end{tabularx}
\end{tcolorbox}

\begin{tcolorbox}[breakable,colback=gray!2,colframe=gray!50,title={Example 4}]
\footnotesize
\begin{tabularx}{\linewidth}{>{\raggedright\arraybackslash\bfseries}p{0.27\linewidth}X}
Edit query & What band is represented in this image? \\
Text generalization query & Which musical group is depicted in the picture? \\
Original entity & Bee Gees \\
Target entity & The Clash \\
EIC question & What band is Bee Gees? \\
EIC target answer & The Clash \\
Image locality question & Who is the person present in this image? \\
Image locality answer & Alan Parsons \\
NBG question & \textbf{1-hop}: What music genre is associated with the group shown in the picture? \\
NBG target answer & Punk rock \\
OBP question & \begin{minipage}[t]{\linewidth}\raggedright the group in this image is a music group from\\
A. australia\\
B. england\\
Answer with one letter only (A or B):\end{minipage} \\
OBP old-binding answer & A \\
\end{tabularx}
\end{tcolorbox}

\begin{tcolorbox}[breakable,colback=gray!2,colframe=gray!50,title={Example 5}]
\footnotesize
\begin{tabularx}{\linewidth}{>{\raggedright\arraybackslash\bfseries}p{0.27\linewidth}X}
Edit query & What country's flag is seen in the image? \\
Text generalization query & Which Asian country is represented in the picture? \\
Original entity & Belgium \\
Target entity & Japan \\
EIC question & What country's flag is Belgium? \\
EIC target answer & Japan \\
Image locality question & Which noble house is depicted in the image? \\
Image locality answer & House of Oldenburg \\
NBG question & \textbf{1-hop}: What is the capital city of the country associated with the item in the picture? \\
NBG target answer & Tokyo \\
OBP question & \begin{minipage}[t]{\linewidth}\raggedright the country in this image is a country in\\
A. europe\\
B. asia\\
Answer with one letter only (A or B):\end{minipage} \\
OBP old-binding answer & A \\
\end{tabularx}
\end{tcolorbox}

\subsection{Details on Metrics}
\label{apd:metrics}

In this subsection, we detail the computation rules for EC-Bench metrics.

We compute EC-Bench metrics with a common token-level scoring rule whenever a task has a prompt and a reference answer. Let $\tau$ denote an evaluation task and let the $j$-th evaluation sample in EC-Bench be:
\begin{equation}
u_j^\tau=(x_j^\tau, v_j^\tau), \qquad
\mathbf{y}_j^\tau=(y_{j,1}^\tau,\ldots,y_{j,L_j}^\tau),
\end{equation}
where $x_j^\tau$ is the text prompt, $v_j^\tau$ is the optional image input, and $\mathbf{y}_j^\tau$ is the non-padding token sequence of the reference answer. We use $s\in\{\mathrm{pre},\mathrm{post}\}$ to denote the model state before and after editing. Given state $s$, the model $f^{(s)}$ generates a next token distribution
\begin{equation}
p_{j,t}^{(s)}(w)=p_{f^{(s)}}\!\left(w \mid u_j^\tau, y_{j,<t}^\tau\right).
\end{equation}
The corresponding predicted token is
\begin{equation}
\hat{y}_{j,t}^{(s)}=\arg\max_w p_{j,t}^{(s)}(w).
\end{equation}
The accuracy for the $j$-th sample and its probability score are then computed as
\begin{equation}
\operatorname{acc}_{j}^{(s)}
=\frac{1}{L_j}\sum_{t=1}^{L_j}
\mathbf{1}\!\left[\hat{y}_{j,t}^{(s)}=y_{j,t}^\tau\right],
\qquad
\operatorname{prob}_{j}^{(s)}
=\frac{1}{L_j}\sum_{t=1}^{L_j}
p_{j,t}^{(s)}\!\left(y_{j,t}^\tau\right).
\end{equation}
The reported task-level scores are averages over the valid evaluation samples:
\begin{equation}
\operatorname{Acc}_{\tau}^{(s)}
=\frac{1}{N_\tau}\sum_{j=1}^{N_\tau}\operatorname{acc}_{j}^{(s)},
\qquad
\operatorname{Prob}_{\tau}^{(s)}
=\frac{1}{N_\tau}\sum_{j=1}^{N_\tau}\operatorname{prob}_{j}^{(s)}.
\end{equation}
Thus, the probability score is an average token probability, rather than the product probability of the whole answer sequence. In the main result tables, edited rows report post-edit scores unless explicitly stated otherwise, while the \texttt{base (unedited)} rows report pre-edit scores.

In practice, the reference answer tokens depend on the task. For efficacy, text generalization, and image generalization, $\mathbf{y}_j^\tau$ is the target-side answer associated with the edit, typically the target entity $e^*$. For EIC, the reference answer is $e^*$; For OBP, the reference answer is the option or answer related to original-entity in the question; For NBG, the reference answer is the target-side associated fact in the open-ended image-conditioned query.

To be noted, in the main text we use the Acc metric for analysis. We additionally provide calculation results based on the probability metric in Appendix.\ref{apd:prob-expr} as supplementary reference.

\paragraph{Locality Metrics.}
Locality metrics use a different rule because their goal is not to reward a new target answer, but to measure whether unrelated behavior is preserved. For a locality sample, let $\hat{\mathbf{y}}_{j,\mathrm{loc}}^{(\mathrm{pre})}$ and $\hat{\mathbf{y}}_{j,\mathrm{loc}}^{(\mathrm{post})}$ denote the predicted tokens, or the selected prediction identifiers, produced by the pre- and post-edit models on the same locality input. The locality score is computed as a consistency rate:
\begin{equation}
\operatorname{Loc}_{\tau}
=\frac{1}{N_\tau}\sum_{j=1}^{N_\tau}
\frac{1}{M_j}\sum_{t=1}^{M_j}
\mathbf{1}\!\left[
\hat{y}_{j,\mathrm{loc},t}^{(\mathrm{pre})}
=
\hat{y}_{j,\mathrm{loc},t}^{(\mathrm{post})}
\right].
\end{equation}
This rule is used for both text locality and multimodal locality, with the latter applying the same consistency principle to image-conditioned locality inputs.

\section{Experiment Setup Details}
\label{apd:setup}

Our experiments build on the codebase implemented by \citet{NEURIPS2024VLKEB}. All the baseline implementations, including
hyperparameters, remain consistent with the setup of \citet{NEURIPS2024VLKEB}.

\subsection{Baselines}
\label{apd:baselines}

We focus on six representative editing methods: FT-LLM, FT-Vis, KE, MEND, IKE, and SERAC, spanning four broad paradigms.
\begin{itemize}
\item \textbf{FT} directly fine-tunes different components of the LVLM. It contains two variants: \textbf{FT-LLM} fine-tunes the LLM backbone, while \textbf{FT-Vis} fine-tunes the vision encoder module or projector.
\item \textbf{KE} \citep{de-cao-etal-2021-editing} is a hypernetwork-based editing method that trains a bidirectional LSTM hypernetwork to predict weight updates to specific layers of the LLM directly based on gradients.
\item \textbf{MEND} \citep{mitchell2022fast} likewise trains a hypernetwork, but predicts low-rank weight updates to specific LLM layers given the gradient information of an edit pair.
\item \textbf{IKE} \citep{zheng-etal-2023-edit} directly leverages in-context learning to achieve the editing effect, which prepends retrieved demonstration examples to the query context without any parameter modification.
\item \textbf{SERAC} \citep{mitchell2022memory} performs editing via an external memory module, stores edit tuples in an external memory and routes queries through a scope classifier at inference time, leaving base model parameters unchanged.
\end{itemize}

\subsection{Details on EC-Bench Evaluation}

In this section, we briefly introduces the configurations used to obtain the EC-Bench results.

\subsubsection{Training}

\textsc{MEND}, \textsc{SERAC}, and \textsc{KE} require training before evaluation. The trainable editors are trained on 5000 edit cases, and a held-out validation set is used to monitor generalization and select the final checkpoint. 

Table~\ref{tab:app-training-config} groups the shared training settings, \textsc{KE}-specific optimization settings, and model-specific training settings. 

\begin{table}[htbp!]
\centering
\scriptsize
\caption{Training configuration for trained editors.}
\label{tab:app-training-config}
\begin{minipage}[htbp!]{0.47\linewidth}
\centering
\textbf{Shared settings.}
\vspace{0.25em}
\begin{tabularx}{\linewidth}{@{}>{\raggedright\arraybackslash}p{0.48\linewidth}X@{}}
\toprule
Setting & Value \\
\midrule
Training cases & 5000 \\
Batch size & 1 \\
Validation batch size & 1 \\
Optimizer & Adam \\
Gradient clipping & 100 \\
$c_{\text{edit}}$ & 0.1 \\
$i_{\text{edit}}$ & 0.1 \\
$c_{\text{loc}}$ & 1.0 \\
$c_{\text{base}}$ & 1.0 \\
\bottomrule
\end{tabularx}
\end{minipage}
\hfill
\begin{minipage}[htbp!]{0.47\linewidth}
\centering
\textbf{\textsc{KE}-specific settings.}
\vspace{0.25em}
\begin{tabularx}{\linewidth}{@{}>{\raggedright\arraybackslash}p{0.48\linewidth}X@{}}
\toprule
Setting & Value \\
\midrule
$lr$ & $3\times10^{-4}$ \\
$lr_{\alpha}$ & $1\times10^{-1}$ \\
Weight decay & 0.01 \\
Total updates & 30k \\
Warmup updates & 1k \\
Maximum length & 32 \\
$\epsilon$ & 0.1 \\
Divergence & $L_p$, $p=2$ \\
Margin range & $[10^{-7},\,10^{-3}]$ \\
\bottomrule
\end{tabularx}
\end{minipage}
\vspace{0.30em}

\textbf{Model-specific trained-editor settings.}
\vspace{0.25em}

\begingroup
\scriptsize
\setlength{\tabcolsep}{2.4pt}
\renewcommand{\arraystretch}{1.22}
\begin{tabularx}{\textwidth}{@{}>{\centering\arraybackslash}p{0.084\textwidth} >{\centering\arraybackslash}p{0.110\textwidth} >{\raggedright\arraybackslash\hspace{0.30em}}p{0.398\textwidth} >{\centering\arraybackslash}p{0.080\textwidth} >{\centering\arraybackslash}p{0.088\textwidth} >{\centering\arraybackslash}p{0.076\textwidth} >{\centering\arraybackslash}p{0.078\textwidth}@{}}
\toprule
Method & LVLM & \multicolumn{1}{c}{Edit location} & Iterations & Early stop & $lr$ & edit lr \\
\midrule
\multirow[c]{4}{*}{\textsc{MEND}} & LLaVA & layers 29--31 MLP, \texttt{down\_proj}/\texttt{up\_proj} & 50k & 5k & \multirow[c]{4}{*}{1e-6} & \multirow[c]{4}{*}{1e-4} \\
& MiniGPT-4 & layers 29--31 MLP, \texttt{down\_proj}/\texttt{up\_proj} & 50k & 5k & & \\
& Qwen-VL & layers 29--31 MLP, \texttt{w1}/\texttt{w2}/\texttt{c\_proj} & 50k & 20k & & \\
& Owl-2 & layers 29--31 MLP, \texttt{down\_proj}/\texttt{up\_proj} & 60k & 20k & & \\
\midrule
\multirow[c]{4}{*}{\textsc{SERAC}} & LLaVA & layers 29--31 MLP, \texttt{down\_proj}/\texttt{up\_proj} & 50k & 10k & \multirow[c]{4}{*}{1e-5} & \multirow[c]{4}{*}{1e-2} \\
& MiniGPT-4 & layers 29--31 MLP, \texttt{down\_proj}/\texttt{up\_proj} & 50k & 5k & & \\
& Qwen-VL & layers 29--31 MLP, \texttt{w1}/\texttt{w2}/\texttt{c\_proj} & 50k & 10k & & \\
& Owl-2 & layers 29--31 MLP, \texttt{down\_proj}/\texttt{up\_proj} & 50k & 10k & & \\
\midrule
\multirow[c]{4}{*}{\textsc{KE}} & LLaVA & layers 29--31 MLP, \texttt{down\_proj}/\texttt{up\_proj} & \multirow[c]{4}{*}{30k} & \multirow[c]{4}{*}{\textemdash} & \multirow[c]{4}{*}{3e-4} & \multirow[c]{4}{*}{\textemdash} \\
& MiniGPT-4 & layers 29--31 MLP, \texttt{down\_proj}/\texttt{up\_proj} & & & & \\
& Qwen-VL & layers 29--31 MLP, \texttt{w1}/\texttt{w2}/\texttt{c\_proj} & & & & \\
& Owl-2 & layers 29--31 MLP, \texttt{down\_proj}/\texttt{up\_proj} & & & & \\
\bottomrule
\end{tabularx}
\endgroup
\end{table}

The shared panel lists the batch size, optimizer, gradient clipping, and loss weights used by the trained editors. The \textsc{KE}-specific panel records additional objective and optimization parameters used only by \textsc{KE}. In the lower panel, iterations is the training budget, early stop is the patience window used for checkpoint selection, $lr$ is the optimizer learning rate, and edit lr is the update scale used by the editor-specific update mechanism.

The lower panel reports only the method- and LVLM-specific settings that differ across trained editors, while shared values are kept in the upper panels. A dash denotes a parameter that is not used by the corresponding method. 

\textsc{MEND} and \textsc{SERAC} use validation-based early stopping, whereas \textsc{KE} uses a fixed update budget. Validation is run every 1k steps for these trained editors, and the selected checkpoint is the one with the best validation performance. For \textsc{MEND}, the learned learning-rate parameters use $lr_{lr}=1e\text{-}4$.

\subsubsection{Evaluation}

Table~\ref{tab:app-ecbench-test-time} summarizes the test-time configuration used by each method. 

\begin{table}[htbp]
\centering
\small
\caption{Test-time configuration for EC-Bench methods.}
\label{tab:app-ecbench-test-time}
\begingroup
\setlength{\tabcolsep}{4.2pt}
\renewcommand{\arraystretch}{1.20}
\begin{tabularx}{\textwidth}{@{}>{\centering\arraybackslash}p{0.11\textwidth} >{\centering\arraybackslash}p{0.14\textwidth} >{\raggedright\arraybackslash}X >{\centering\arraybackslash}p{0.09\textwidth} >{\centering\arraybackslash}p{0.10\textwidth}@{}}
\toprule
Method & LVLM & Edit location & \# edit steps & edit lr \\
\midrule
\multirow[c]{4}{*}{\textsc{FT}} & LLaVA & layer 31 MLP, \texttt{down\_proj}/\texttt{up\_proj} & 10 & $1\times10^{-4}$ \\
& MiniGPT-4 & layer 31 MLP, \texttt{down\_proj}/\texttt{up\_proj} & 3 & $1\times10^{-4}$ \\
& Qwen-VL & layer 31 MLP, \texttt{w1}/\texttt{w2}/\texttt{c\_proj} & 2 & $1\times10^{-4}$ \\
& Owl-2 & layer 31 MLP, \texttt{gate\_proj}/\texttt{down\_proj}/\texttt{up\_proj} & 20 & $1\times10^{-4}$ \\
\midrule
\multirow[c]{4}{*}{\textsc{FT-VIS}} & LLaVA & multimodal projector, \texttt{mm\_projector} & 10 & $1\times10^{-4}$ \\
& MiniGPT-4 & Q-Former & 15 & $1\times10^{-4}$ \\
& Qwen-VL & final visual-transformer MLP, resblock 47 & 25 & $2\times10^{-3}$ \\
& Owl-2 & vision model & 25 & $1\times10^{-3}$ \\
\midrule
\multirow[c]{4}{*}{\textsc{MEND}} & LLaVA & layers 29--31 MLP, \texttt{down\_proj}/\texttt{up\_proj} & \textemdash & \textemdash \\
& MiniGPT-4 & layers 29--31 MLP, \texttt{down\_proj}/\texttt{up\_proj} & \textemdash & \textemdash \\
& Qwen-VL & layers 29--31, \texttt{w1}/\texttt{w2}/\texttt{c\_proj} & \textemdash & \textemdash \\
& Owl-2 & layers 29--31, \texttt{down\_proj}/\texttt{up\_proj} & \textemdash & \textemdash \\
\midrule
\multirow[c]{4}{*}{\textsc{SERAC}} & LLaVA & layers 29--31 MLP, \texttt{down\_proj}/\texttt{up\_proj} & \textemdash & \textemdash \\
& MiniGPT-4 & layers 29--31 MLP, \texttt{down\_proj}/\texttt{up\_proj} & \textemdash & \textemdash \\
& Qwen-VL & layers 29--31, \texttt{w1}/\texttt{w2}/\texttt{c\_proj} & \textemdash & \textemdash \\
& Owl-2 & layers 29--31, \texttt{down\_proj}/\texttt{up\_proj} & \textemdash & \textemdash \\
\midrule
\multirow[c]{4}{*}{\textsc{KE}} & LLaVA & layers 29--31, \texttt{down\_proj}/\texttt{up\_proj} & \textemdash & \textemdash \\
& MiniGPT-4 & layers 29--31, \texttt{down\_proj}/\texttt{up\_proj} & \textemdash & \textemdash \\
& Qwen-VL & layers 29--31, \texttt{w1}/\texttt{w2}/\texttt{c\_proj} & \textemdash & \textemdash \\
& Owl-2 & layers 29--31, \texttt{down\_proj}/\texttt{up\_proj} & \textemdash & \textemdash \\
\midrule
\textsc{IKE} & all & retrieved demonstrations ($k=32$) with \texttt{all-MiniLM-L6-v2} & \textemdash & \textemdash \\
\bottomrule
\end{tabularx}
\endgroup
\end{table}

The edit location specifies the model component or parameter group to which an edit is applied. For per-case update methods, \# edit steps is the number of gradient-update steps and edit lr is the corresponding learning rate. \textsc{MEND}, \textsc{SERAC}, and \textsc{KE} apply trained editors without additional test-time gradient steps; and \textsc{IKE} is shown separately because it uses retrieved in-context demonstrations rather than an edited parameter location. All methods are evaluated as single-sample edits, where each edit case is handled independently.

\subsection{Details on Editing-Location Control Experiments}

The editing-location control experiments vary the editing location for \textsc{FT} and \textsc{MEND}. Table~\ref{tab:app-editing-location-control-setup} lists the language-side locations included in the comparison and the Vis setting used as the vision-side reference. 

\begin{table}[htbp!]
\centering
\small
\caption{Configuration of the editing-location control experiments.}
\label{tab:app-editing-location-control-setup}
\begingroup
\setlength{\tabcolsep}{2.5pt}
\renewcommand{\arraystretch}{1.40}
\begin{tabularx}{\textwidth}{@{}>{\centering\arraybackslash}p{0.080\textwidth} >{\centering\arraybackslash}p{0.112\textwidth} >{\centering\arraybackslash}p{0.290\textwidth} >{\raggedright\arraybackslash}p{0.310\textwidth} >{\centering\arraybackslash}p{0.085\textwidth} >{\centering\arraybackslash}p{0.060\textwidth}@{}}
\toprule
Method & LVLM & Edit location & Parameter group & \# edit steps & edit lr \\
\midrule
\multirow[c]{4}{*}{\textsc{FT}} & \multirow[c]{2}{*}{LLaVA} & layers 05, 31 MLP & \texttt{down\_proj}/\texttt{up\_proj} & \multirow[c]{2}{*}{10} & \multirow[c]{2}{*}{1e-4} \\
& & multimodal projector & \texttt{mm\_projector} & & \\
\cmidrule(lr){2-6}
& \multirow[c]{2}{*}{MiniGPT-4} & layers 10, 31 MLP & \texttt{down\_proj}/\texttt{up\_proj} & 3 & \multirow[c]{2}{*}{1e-4} \\
& & Q-Former & \textemdash & 15 & \\
\midrule
\multirow[c]{6}{*}{\textsc{MEND}} & \multirow[c]{3}{*}{LLaVA} & layers 15--17 MLP & \texttt{down\_proj}/\texttt{up\_proj} & \textemdash & \textemdash \\
& & layers 29--31 MLP & \texttt{down\_proj}/\texttt{up\_proj} & \textemdash & \textemdash \\
& & multimodal projector & \texttt{mm\_projector} & \textemdash & \textemdash \\
\cmidrule(lr){2-6}
& \multirow[c]{3}{*}{MiniGPT-4} & layers 1--3 MLP & \texttt{down\_proj}/\texttt{up\_proj} & \textemdash & \textemdash \\
& & layers 29--31 MLP & \texttt{down\_proj}/\texttt{up\_proj} & \textemdash & \textemdash \\
& & Q-Former layer 11 & \texttt{intermediate/output\_query} & \textemdash & \textemdash \\
\bottomrule
\end{tabularx}
\endgroup
\end{table}

In this table, edit location specifies either the selected LLM layers or the vision-side setting, parameter group specifies the edited module within that location, and the last two columns use the same per-case update notation as Table~\ref{tab:app-ecbench-test-time}. For \textsc{MEND}, dashes in the last two columns indicate that the trained editor is applied without additional test-time gradient steps. The language-side rows vary the edited LLM MLP layers, while the VIS rows move the editing location to the vision-side component for the corresponding method. In addition, the editing locations of FT-shallow selected in Table \ref{tab:edit-location} are layer 5 and layer 10 for LLaVA and MiniGPT-4, respectively;

\section{Supplementary Experimental Results}

\subsection{Results on Owl-2}
\label{apd:owl-results}

In this section, we extend our experiments to the mPLUG-Owl2 model using the EC-Bench dataset, with results presented in Tables \ref{tab:owl-main}. Consistent with findings in Section \ref{subsec:main-expr}, EIC persists on this model, where most methods also exhibit an increase in EIC, along with anomalies on the OBP and NBG tasks. The conclusions on FT-Vis also apply to this model, where it achieves the best comprehensive performance across these three metrics.

% Auto-generated from whitelist-backed results.
\begin{table*}[ht!]
\centering
\footnotesize
\setlength{\tabcolsep}{3.1pt}
\renewcommand{\arraystretch}{1.10}

\caption{Main EC-Bench results on inherited and diagnostic metrics.}
\label{tab:owl-main} 
\sf
\resizebox{\textwidth}{!}{%

\begin{tabular}{@{}cccccccccc@{}}
\toprule
Model & Method & Efficacy $\uparrow$ & T-Gen $\uparrow$ & I-Gen $\uparrow$ & T-Loc $\uparrow$ & I-Loc $\uparrow$ & EIC $\downarrow$ & OBP $\downarrow$ & NBG $\uparrow$ \\
\midrule

\multirow[c]{7}{*}{\textbf{Owl-2}} & \textit{base (unedited)} & 32.8 & 36.9 & 32.8 & 100.0 & 100.0 & 28.3 & 84.7 & 43.8 \\
\cmidrule(lr){2-10}
 & FT & 100.0 & 99.6 & 100.0 & 57.5 & 41.1 & 98.0 & 83.6 & 37.2 \\
 & FT-VIS & 99.9 & 97.6 & 99.4 & 100.0 & 11.8 & 28.3 & 29.4 & 52.3 \\
% \cmidrule(lr){2-10}
 & MEND & 99.4 & 98.5 & 98.3 & 86.9 & 90.7 & 78.4 & 84.6 & 43.8 \\
 & SERAC & 98.5 & 96.3 & 98.3 & 40.1 & 3.0 & 78.2 & 84.9 & 54.4 \\
 & IKE & 100.0 & 99.6 & 100.0 & 44.6 & 16.5 & 64.8 & 26.7 & 64.8 \\
 & KE & 79.4 & 75.7 & 75.8 & 82.2 & 51.5 & 55.6 & 86.1 & 46.8 \\
\bottomrule
\end{tabular}}

% \vspace{-2em}
\end{table*}

\subsection{Probability Metric Results on Ec-Bench}
\label{apd:prob-expr}

In addition to the accuracy metrics reported in the main text, we additionally provide metric results based on probability computation (See Appendix.\ref{apd:metrics}) as a supplement. The probability-based metrics, being continuous metrics, offer finer granularity and are provided for reference. Table~\ref{tab:app-ecbench-prob} reports the EC-Bench main experiment results under the probability metric. Note that, since probability computation requires specifying the target answer token sequence, and the locality metric has no target answer as it only evaluates the consistency of outputs before and after editing, the locality results are not reported here.

% Auto-generated from whitelist-backed results.
\begin{table}[t]
\centering
\footnotesize
\setlength{\tabcolsep}{3.4pt}
\renewcommand{\arraystretch}{1.10}
\caption{Probability results for the main EC-Bench results.}
\label{tab:app-ecbench-prob}
\resizebox{\textwidth}{!}{%
\begin{tabular}{@{}cccccccc@{}}
\toprule
Model & Method & Efficacy $\uparrow$ & T-Gen $\uparrow$ & I-Gen $\uparrow$ & EIC $\downarrow$ & OBP $\downarrow$ & NBG $\uparrow$ \\
\midrule
\multirow[c]{7}{*}{\textbf{LLaVA}} & \textit{base (unedited)} & 26.1 & 29.1 & 25.8 & 24.0 & 88.1 & 32.9 \\
\cmidrule(lr){2-8}
 & FT & 99.4 & 98.9 & 99.4 & 98.8 & 68.6 & 29.3 \\
 & FT-VIS & 98.5 & 91.7 & 87.9 & 24.0 & 50.9 & 45.0 \\
\cmidrule(lr){2-8}
 & MEND & 98.5 & 98.2 & 98.4 & 96.3 & 88.0 & 36.7 \\
 & SERAC & 98.8 & 96.9 & 98.8 & 74.4 & 87.6 & 44.7 \\
 & IKE & 99.6 & 97.9 & 99.6 & 65.9 & 47.1 & 51.0 \\
 & KE & 97.8 & 96.9 & 97.6 & 92.1 & 88.0 & 35.9 \\
\midrule
\multirow[c]{7}{*}{\textbf{MiniGPT-4}} & \textit{base (unedited)} & 22.9 & 25.7 & 22.7 & 27.3 & 49.8 & 31.5 \\
\cmidrule(lr){2-8}
 & FT & 98.6 & 97.6 & 97.6 & 66.4 & 41.8 & 31.9 \\
 & FT-VIS & 99.9 & 98.7 & 99.6 & 27.3 & 50.9 & 35.7 \\
\cmidrule(lr){2-8}
 & MEND & 99.0 & 98.6 & 98.8 & 91.9 & 51.1 & 35.6 \\
 & SERAC & 97.3 & 93.5 & 97.3 & 75.9 & 49.6 & 46.2 \\
 & IKE & 99.0 & 97.1 & 99.0 & 67.2 & 42.6 & 46.9 \\
 & KE & 97.1 & 96.8 & 96.9 & 80.4 & 25.3 & 34.5 \\
\midrule
\multirow[c]{7}{*}{\textbf{Qwen-VL}} & \textit{base (unedited)} & 20.8 & 24.0 & 20.7 & 19.1 & 56.6 & 25.9 \\
\cmidrule(lr){2-8}
 & FT & 99.8 & 96.9 & 99.4 & 82.7 & 38.7 & 25.4 \\
 & FT-VIS & 100.0 & 93.1 & 99.3 & 19.1 & 29.9 & 27.7 \\
\cmidrule(lr){2-8}
 & MEND & 99.3 & 98.1 & 97.4 & 63.4 & 57.8 & 28.8 \\
 & SERAC & 66.7 & 62.6 & 66.8 & 50.4 & 45.0 & 25.4 \\
 & IKE & 99.2 & 97.8 & 99.2 & 55.7 & 34.1 & 46.9 \\
 & KE & 98.8 & 95.0 & 98.2 & 88.3 & 37.9 & 29.1 \\
\midrule
\multirow[c]{7}{*}{\textbf{Owl-2}} & \textit{base (unedited)} & 28.1 & 32.0 & 28.1 & 24.4 & 82.1 & 37.9 \\
\cmidrule(lr){2-8}
 & FT & 100.0 & 99.5 & 100.0 & 97.0 & 80.9 & 33.1 \\
 & FT-VIS & 99.7 & 95.4 & 99.0 & 24.4 & 30.5 & 46.4 \\
\cmidrule(lr){2-8}
 & MEND & 99.1 & 98.0 & 97.9 & 75.9 & 82.0 & 38.9 \\
 & SERAC & 97.7 & 94.4 & 97.6 & 76.5 & 81.2 & 49.9 \\
 & IKE & 99.8 & 98.7 & 99.8 & 63.6 & 28.1 & 54.7 \\
 & KE & 63.6 & 62.5 & 62.1 & 47.8 & 83.7 & 40.2 \\
\bottomrule
\end{tabular}%
}
\end{table}

Table~\ref{tab:edit-location-prob} provides the results of the editing-location control experiment under the probability metric. All other experimental settings are identical to those in the main text.

% Auto-generated from whitelist-backed results.
\begin{table*}[htbp]
\centering

\footnotesize
\setlength{\tabcolsep}{2.8pt}
\renewcommand{\arraystretch}{1.06}

\caption{Editing-location comparison for FT and MEND.}
\label{tab:edit-location-prob}
\sf
\resizebox{\textwidth}{!}{%

\begin{tabular}{llcccccc}
\toprule
Model & Method & Efficacy $\uparrow$ & T-Gen $\uparrow$ & I-Gen $\uparrow$ & EIC $\downarrow$ & OBP $\downarrow$ & NBG $\uparrow$ \\

\midrule

\multirow[c]{8}{*}{LLaVA}
& base (unedited) & 26.1 & 29.1 & 25.8 & 24.0 & 88.1 & 32.9 \\
\cmidrule(lr){2-8}
& FT-Shallow & 98.0 & 82.9 & 92.5 & 39.2 & 67.3 & 43.8 \\
& FT-Deep & 99.4 & 98.9 & 99.4 & 98.8 & 68.6 & 29.3 \\
& FT-Vis & 98.5 & 91.7 & 87.9 & 24.0 & 50.9 & 45.0 \\
\cmidrule(lr){2-8}
& MEND-Shallow & 97.7 & 96.1 & 96.5 & 56.7 & 88.9 & 37.4 \\
& MEND-Deep & 98.5 & 98.2 & 98.4 & 96.3 & 88.0 & 36.7 \\
& MEND-Vis & 60.0 & 60.8 & 58.3 & 24.0 & 83.2 & 45.4 \\

\midrule
\midrule

\multirow[c]{8}{*}{MiniGPT-4}
& base (unedited) & 22.9 & 25.7 & 22.7 & 27.3 & 49.8 & 31.5 \\
\cmidrule(lr){2-8}
& FT-Shallow & 66.8 & 60.5 & 62.6 & 37.8 & 51.7 & 38.2 \\
& FT-Deep & 98.6 & 97.6 & 97.6 & 66.4 & 41.8 & 31.9 \\
& FT-Vis & 99.9 & 98.7 & 99.6 & 27.3 & 50.9 & 35.7 \\
\cmidrule(lr){2-8}
& MEND-Shallow & 83.6 & 84.1 & 82.5 & 37.6 & 49.5 & 38.6 \\
& MEND-Deep & 99.0 & 98.6 & 98.8 & 91.9 & 51.1 & 35.6 \\
& MEND-Vis & 81.9 & 81.2 & 69.6 & 27.3 & 45.5 & 37.8 \\
\bottomrule

\end{tabular}}

\vspace{-2em}
\end{table*}

%%%%%%%%%%%%%%%%%%%%%%%%%%%%%%%%%%%%%%%%%%%%%%%%%%%%%%%%%%%%

\clearpage
\end{document}